\documentclass[11pt,a4paper]{article}
\usepackage[hyperref]{emnlp2018}

\usepackage{graphicx}
\usepackage{times}
\usepackage{latexsym}
\usepackage{caption}
\usepackage{url}
\usepackage{amsmath}
\usepackage{amsfonts}
\usepackage{amssymb}
\usepackage{multirow}
\usepackage{arydshln}
\usepackage{nicefrac}
\usepackage{subfig}
\usepackage{float}

\aclfinalcopy 

\makeatletter
\newcommand{\printfnsymbol}[1]{%
  \textsuperscript{\@fnsymbol{#1}}%
}
\makeatother

\title{Unsupervised Word Polysemy Quantification with Multiresolution Grids of Contextual Embeddings}

\author{Christos Xypolopoulos\thanks{~~Equal contribution. CX handled the data, generated the rankings, and sampled the examples of section \ref{sec:app}. AJPT computed the results, plots, and wrote the paper. Both authors participated in the design of the study.} \\
{\'Ecole Polytechnique}
\And
  Antoine J.-P. Tixier\printfnsymbol{1} \\
  {\'Ecole Polytechnique} \And
  Michalis Vazirgiannis\\
  {\'Ecole Polytechnique \& AUEB}}

\date{}

\begin{document}
\maketitle

\begin{abstract}
The number of senses of a given word, or polysemy, is a very subjective notion, which varies widely across annotators and resources.
In this paper\footnote{Accepted as a long paper in EACL 2021.}, we propose a novel method to estimate polysemy based on simple geometry in the contextual embedding space. 
Our approach is fully unsupervised and purely data-driven.
Through rigorous experiments, we show that our rankings are well correlated, with strong statistical significance, with 6 different rankings derived from famous human-constructed resources such as WordNet, OntoNotes, Oxford, Wikipedia, etc., for 6 different standard metrics.
We also visualize and analyze the correlation between the human rankings and make interesting observations.
A valuable by-product of our method is the ability to sample, at no extra cost, sentences containing different senses of a given word.
Finally, the fully unsupervised nature of our approach makes it applicable to any language.
Code and data are publicly available\footnote{https://github.com/ksipos/polysemy-assessment}.
\end{abstract}

\begin{figure*}
\centering
\includegraphics[width=0.85\textwidth]{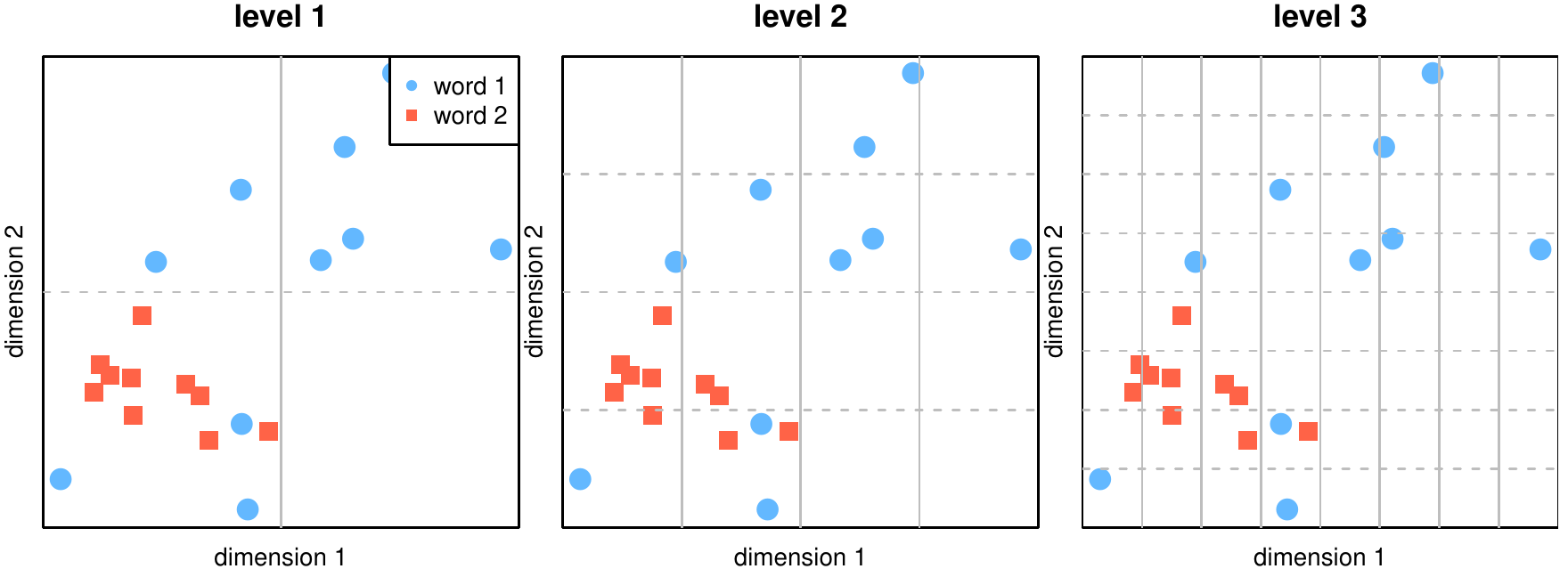}
\captionsetup{size=small}
\caption{Illustration of the proposed approach with $D=2$ and $L=3$. \label{fig:demo}}
\end{figure*}

\section{Introduction}
Polysemy, the number of senses that a word has, is a very subjective notion, subject to individual biases.
Word sense annotation has always been one of the tasks with the lowest values of inter-annotator agreement \citep{artstein2008inter}.
Yet, creating high-quality, consistent word sense inventories is a critical pre-requisite to successful word sense disambiguation.

Towards creating word sense inventories, it can be helpful to have some reliable information about polysemy.
That is, knowing which words have many senses and which words have only a few senses.
Such information can help in creating new inventories but also in validating and interpreting existing ones.
It can also help select which words to include in a study (e.g., only highly polysemous words).

We propose a novel, fully unsupervised, and data-driven approach to quantify polysemy, based on basic geometry in the contextual embedding space.

Contextual word embeddings have emerged in the last few years as part of the NLP transfer learning revolution.
Now, entire deep models are pre-trained on huge amounts of unannotated data and fine-tuned on much smaller annotated datasets.
Some of the most famous examples include ULMFiT \citep{howard2018universal} and ELMo \citep{peters2018deep}, both based on recurrent neural networks; and GPT \citep{radford2018improving} and BERT \citep{devlin2018bert}, based on transformers \citep{vaswani2017attention}.
These models all are very deep \textit{language models}.
During pre-training on large-scale corpora, they learn to generate powerful internal representations, including fine-grained contextual word embeddings.
For instance, in a well pre-trained model, the word \textit{python} will have two very different embeddings depending on whether it occurs in a programming context (as in, e.g., ``I love to write code in python'') or in an ecological context (``while hiking in the rainforest, I saw a python'').

Our approach capitalizes on the contextual embeddings previously described.
It does not involve any tool and does not rely on any human input or judgment.
Also, thanks to its unsupervised nature, it can be applied to any language (even those with limited resources), provided that contextual embeddings are available. 

The remainder of this paper is organized as follows.
We detail our approach in section \ref{sec:ours}.
Then, we present our experimental setup (sec. \ref{sec:exps}), evaluation metrics (sec. \ref{sec:metrics}), and report and interpret our results (sec. \ref{sec:results}).
In section \ref{sec:app}, we briefly touch on two other interesting applications of our method. One that allows the user to sample sentences containing different senses of a given word and one that goes towards word sense induction.
Finally, related work is presented in section \ref{sec:related}.

\section{Proposed approach}\label{sec:ours}

\subsection{Basic assumption}

First, by passing diverse sentences containing a given word to a pre-trained language model, we construct a representative set of vectors for that word (one vector for each occurrence of the word).
The basic and intuitive assumption we make is that \textit{the volume covered by the cloud of points in the contextual embedding space is representative of the polysemy of the associated word}.

\subsection{Main idea: multiresolution grids}
As a proxy for the volume covered, we adopt a simple geometrical approach.
As shown in Fig. \ref{fig:demo}, we construct a hierarchical discretization of the space, where, at each level, the same number of bins are drawn along each dimension.
Each level corresponds to a different resolution.
Our polysemy score is based on the proportion of bins covered by the vectors of a given word at each level.

\noindent \textbf{Grid vs. clustering}. Using a binning strategy makes more sense than a clustering-based approach.
Indeed, clusters do not partition the space equally and regularly.
This is especially problematic since word representations are not uniformly distributed in the embedding space \citep{ethayarajh2019contextual}.
Therefore, the vectors lying in the same dense area of the space will always belong to one single large cluster, while outliers lying in the same, but sparser, area of the space, will be assigned to many different small clusters.
Therefore, counting the number of clusters a given word belongs to is not a reliable indicator of how much of the space this word covers.

\subsection{Scoring scheme}
We quantify the polysemy degree of a word $w$ as:

\begin{equation}\label{eq:scoring}
    \text{score}(w)=\sum_{l=1}^{L}{\frac{\text{coverage}_w^l}{2^{L-l}}}
\end{equation}

\noindent where $\text{coverage}_w^l$ designates the proportion of bins covered by word $w$ at level $l$, between 0 and 1.
At each level, $2^l$ bins are drawn along each dimension (see the vertical and horizontal lines in Fig. \ref{fig:demo}).
The hierarchy starts at $l=1$ since there is only one bin covering all the space at $l=0$ (so all words have equal coverage at this level).
The total number of bins in the entire space, at a given level $l$, is equal to $(2^l)^D$.

Consider again the example of Fig. \ref{fig:demo}.
In this example, each word is associated with a set of 10 contextualized embeddings in a space of dimension $D=2$, and the hierarchy has $L=3$ levels.
First, we can clearly see that word 1 (blue circles) covers a large area of the space while all the vectors of word 2 (orange squares) are grouped in the same region.
Intuitively, this can be interpreted as ``word 1 occurs in more different contexts than word 2'', which per our assumption, is equivalent to saying that ``word 1 is more polysemous than word 2''.

Let us now see how this is reflected by our scoring scheme.
First, the penalization terms (denominators) for levels 1 to 3 are $ \big[\frac{1}{2^2},\frac{1}{2^1},\frac{1}{2^0}\big] = [\frac{1}{4},\frac{1}{2},1\big]$.
Note that the higher the level, the exponentially more bins, and so the less penalized (or the more rewarded) coverage is, because getting good coverage becomes more and more difficult.
Now, per Eq. \ref{eq:scoring}, the score of word 1 is computed as the dot product of its coverage vector $\big[ \frac{3}{4}, \frac{7}{16}, \frac{10}{64} \big]$ (coverage at each level) with the penalization vector, which gives a score of $0.5625$.
Likewise, the score of word 2 is computed as $[\frac{1}{4},\frac{1}{2},1\big] \cdot \big[\frac{1}{4},\frac{4}{16},\frac{7}{64}\big] = 0.297$.
We can thus see that our scores reflect what can be observed in Fig. \ref{fig:demo}: word 1 covers a larger area of the space than word 2.

Note that the score of a given word is only meaningful compared to the scores of other words, i.e., in rankings, as will be seen in the next section.

\section{Experiments}\label{sec:exps}
In this section, we describe the protocol we followed to test the extent to which our rankings match human rankings.

\subsection{Word selection}\label{sub:wordsel}
The first step was to select words to include in our analysis.
For this purpose, we downloaded and extracted all the text from the latest available English Wikipedia dump\footnote{\url{https://dumps.wikimedia.org/}}.
We then performed tokenization, stopword, punctuation, and number removal and counted the occurrence of each token of at least 3 characters in size.
Out of these tokens, we kept the 2000 most frequent.

\subsection{Generating vector sets}
For each word in the shortlist, we randomly selected 3000 sentences such that the corresponding word appeared exactly once within each sentence.
The words that did not appear in at least 3000 sentences were removed from the analysis, reducing the shortlist's size from 2000 to 1822.
Then, for each word, the associated sentences were passed through a pre-trained ELMo model\footnote{We used the implementation and pre-trained weights publicly released by the authors \url{https://allennlp.org/elmo}.} \citep{peters2018deep} in test mode, and the top layer representations corresponding to the word were harvested.
The advantage of using ELMo's top layer embeddings is that they are the most contextual, as shown by \citet{ethayarajh2019contextual}.
We ended up with a set of exactly 3000 1024-dimensional contextual embeddings for each word.

\subsection{Dimensionality reduction}
Remember that the total number of bins in the entire space is equal to $(2^l)^D$ at a given level $l$, which would have given us an infinite number of bins even at the first level, since the ELMo representations have dimensionality $D=1024$.
To reduce the dimensionality of the contextual embedding space, we applied PCA, trying 19 different output dimensionalities, from $2$ to $20$ with steps of $1$.
Due to the quantity and high initial dimensionality of the vectors, we used the distributed\footnote{15 executors with 10 GB of RAM each.} version of PCA provided by the PySpark's ML Library \citep{meng2016mllib}.

\subsection{Score computation}
We computed our scores for each PCA output dimensionality, trying with 18 different hierarchies whose numbers of levels $L$ ranged from 2 to 19.
So in total, we obtained $19 \times 18= 342$ rankings.

\subsection{Ground truth rankings and baselines}
We evaluated the rankings generated by our approach against several ground truth rankings that we derived from human-constructed resources.

Since the number of senses of a word is a subjective, debatable notion, and thus may vary from source to source, we included 6 ground truth rankings in our analysis, in order to minimize source-specific bias as much as possible.
For sanity checking purposes, we also added two basic baseline rankings (frequency and random).
We provide more details about all rankings in what follows.

\subsubsection{WordNet}
We used WordNet \citep{miller1998wordnet} version 3.0 and counted the number of synonym sets or ``synsets'' of each word.

\subsubsection{WordNet-Reduced}
There are very subtle differences among the WordNet senses (``synsets''), making distinguishing between them difficult and even irrelevant in some applications \citep{palmer2004different,palmer2007making,brown2010number,rumshisky2011crowdsourcing,jurgens2013embracing}.
For instance, \textit{call} has 41 senses in the original WordNet (28 as verb and 13 as noun).
Even for other words with fewer senses, like \textit{eating} (7 senses in total), the difference between senses can be very tiny.
For instance, ``take in solid food'' and ``eat a meal; take a meal'' are really close in meaning.
This very fine granularity of WordNet may somewhat artificially increase the polysemy of some words.

To reduce the granularity of the WordNet synsets, we used their sense keys\footnote{See `Sense Key Encoding' here: \url{https://wordnet.princeton.edu/documentation/senseidx5wn}}.
They follow the format {\small \texttt{lemma\%ss\_type:lex\_filenum: lex\_id:head\_word:head\_id}}, where {\small \texttt{ss\_type}} represents the synset type (part-of-speech tag such as noun, verb, adjective) and {\small \texttt{lex\_filenum}} represents the name of the lexicographer file containing the synset for the sense ({\small \texttt{noun.animal}, \texttt{noun.event}, \texttt{verb.emotion}}, etc.).
We truncated the sense keys after {\small \texttt{lex\_filenum}}.

For instance, ``take in solid food'' and ``eat a meal; take a meal'' initially correspond to two different senses with keys {\small \texttt{eat\%2:34:00::}} and {\small \texttt{eat\%2:34:01::}}, but after truncation, they both are mapped to the same sense: {\small \texttt{eat\%2:34}}.
However, coarse differences in senses are still captured. For instance, \textit{bank} ``sloping land'' ({\small \texttt{bank\%1:17:01::}}) and \textit{bank} ``financial institution'' ({\small \texttt{bank\%1:14:00::}}) are still mapped to two different senses after truncation, respectively {\small \texttt{bank\%1:17}} and {\small \texttt{bank\%1:14}}.

\subsubsection{WordNet-Domains}
WordNet Domains \citep{bentivogli2004revising,magnini2000integrating} is a lexical resource created in a semi-automatic way to augment WordNet with domain labels.
Instead of synsets, each word is associated with a number of semantic domains.
The domains are areas of human knowledge (politics, economy, sports, etc.) exhibiting specific terminology and lexical coherence. 
As for the two previous WordNet ground truth rankings, we simply counted the number of domains associated with each word.

\subsubsection{OntoNotes}
OntoNotes \citep{hovy2006ontonotes,weischedel2011ontonotes} is a large annotated corpus comprising various text genres (news, conversational telephone speech, weblogs, newsgroups, broadcast, talk shows) with structural information and shallow semantics.

We counted the senses in the sense inventory of each word.
The senses in OntoNotes are groupings of the WordNet synsets, constructed by human annotators.
As a result, the sense granularity of OntoNotes is coarser than that of WordNet \citep{brown2010number}.

\subsubsection{Oxford}
We counted the number of senses returned by the Oxford dictionary\footnote{\url{www.lexico.com}}, which was, at the time of this study, the resource underlying the Google dictionary functionality.

\subsubsection{Wikipedia}
We capitalized on the Wikipedia disambiguation pages\footnote{www.wikipedia.org/wiki/word\_(disambiguation)}.
Such pages contain a list of the different categories under which one or more articles about the query word can be found.
For example, the disambiguation page of the word \textit{bank} includes categories such as geography, finance, computing (data bank), and science (blood bank). 
We counted the number of categories on the disambiguation page of each word to generate the ranking.

\subsubsection{Frequency and random baselines}
In the frequency baseline, we ranked words in decreasing order of their frequency in the entire Wikipedia dump (see subsection \ref{sub:wordsel}).
The naive assumption made here is that words occurring the most have the most senses.

\begin{figure}
\centering
\includegraphics[width=0.6\columnwidth]{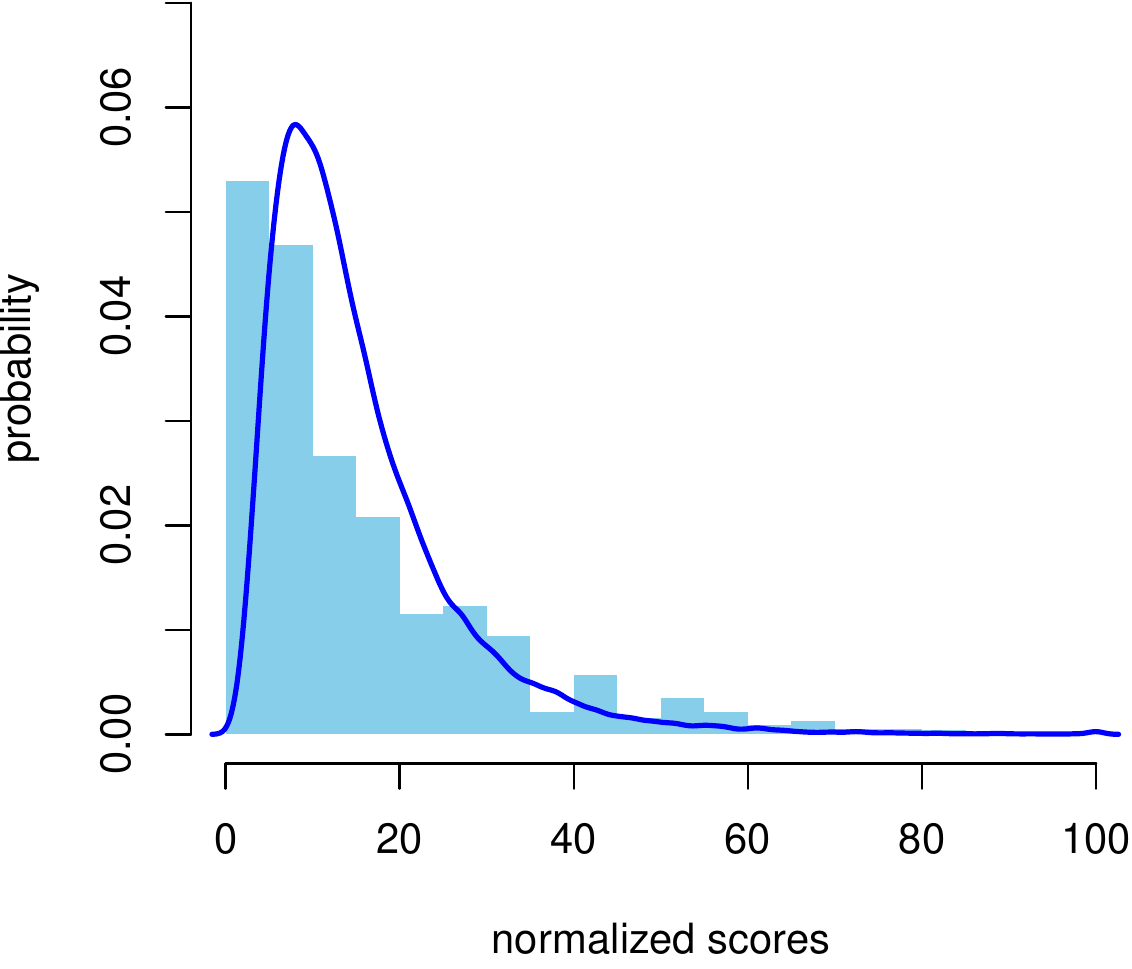}
\captionsetup{size=small}
\caption{Average score distribution of the 5 ground truth rankings and frequency baseline (histogram) vs. average score distribution of the random baseline (\textcolor{blue}{\textbf{blue}} curve). \label{fig:random}}
\end{figure}

\begin{figure*}
\centering
\includegraphics[width=0.725\textwidth]{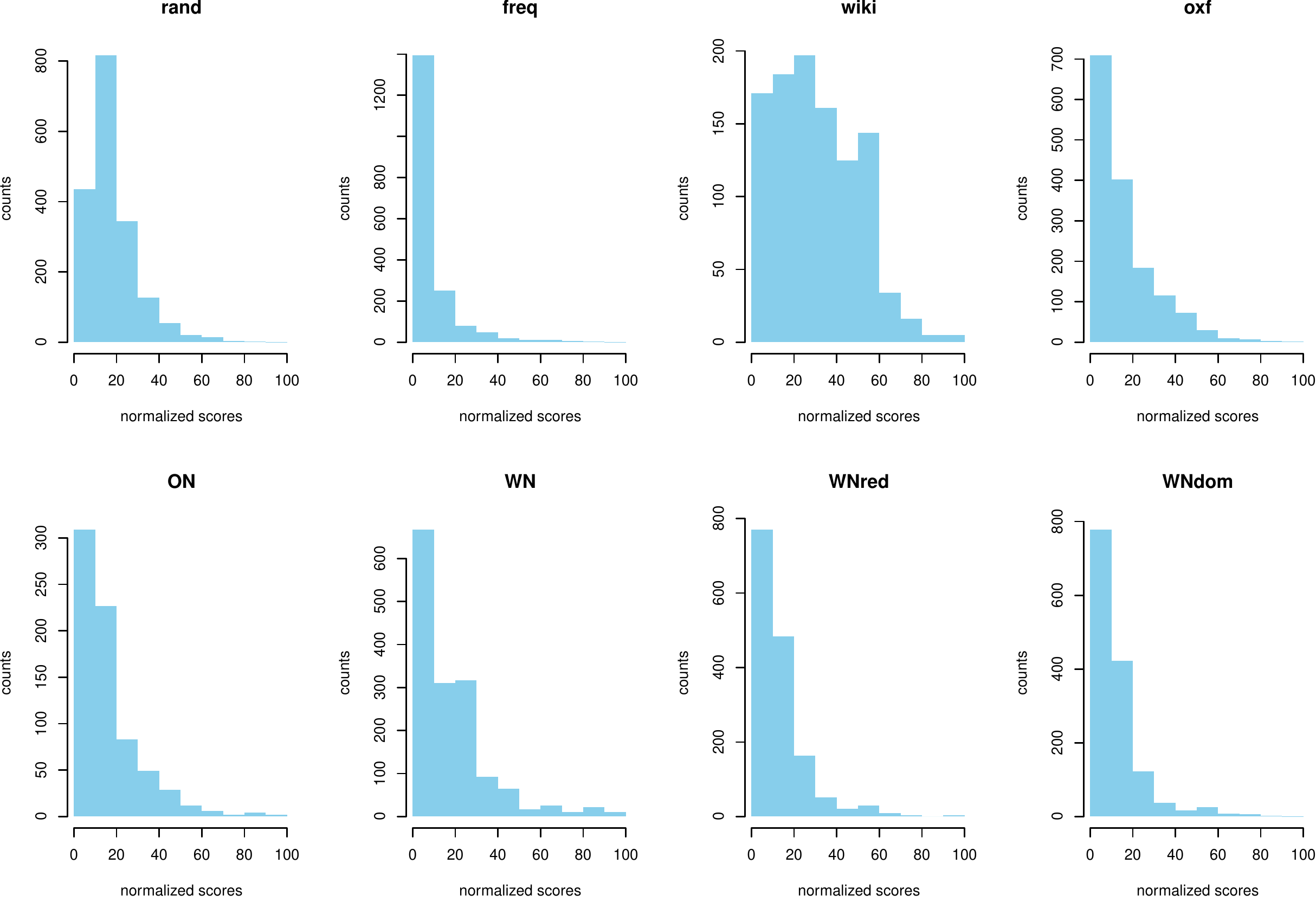}
\captionsetup{size=small}
\caption{Normalized ranking score distributions for the random and frequency rankings and the human-derived ground truth rankings. \label{fig:score_distrs}}
\end{figure*}

With the random baseline, on the other hand, we produced rankings by shuffling words.
Further, we assigned them random scores by sampling from the Log Normal distribution\footnote{with mean and standard deviation 0 and 0.6 (resp.)}, to imitate the long-tail behavior of the other score distributions, as can be seen in Fig. \ref{fig:random}.
All distributions can be seen in Fig. \ref{fig:score_distrs}.
Note that to account for randomness, all results for the random baseline are averages over 30 runs.

\noindent Not every of the 1822 words included in our analysis had an entry in each of the resources described above.
The lengths of each ground truth ranking are shown in Table \ref{table:stats}.

\begin{table}
\center
\scalebox{0.8}{
\begin{tabular}{ll}
\hline
Ranking & \# words \\ \hline
WN & 1535 \\
WN-reduced & 1535 \\
WN-Domains & 1420 \\ 
Oxford & 1536 \\
Wikipedia & 1042 \\
OntoNotes & 723 \\ \hdashline
Frequency \& random & 1822 \\
\end{tabular}
}
\captionsetup{size=small}
\caption{Length of the ground truth rankings. \label{table:stats}}
\end{table}

\section{Evaluation metrics}\label{sec:metrics}
As will be detailed next, we used 6 standard metrics from the fields of statistics and information retrieval to compare among methods.
To ensure fair comparison, the scores in the rankings of all methods were normalized to be in the $[0,100]$ range before proceeding.

Also, each method played in turn the role of candidate and ground truth.
This allowed us to compute not only the similarity between our rankings and the ground truth rankings, but also the similarity among the ground truth rankings themselves, which was interesting for exploration purposes.

For each pair of evaluated and ground truth method, only the parts of the rankings corresponding to the words in common (intersection) were compared.
Thus, the rankings in each (candidate,ground truth) pair had equal length.

\subsection{Similarity and correlation metrics}

\subsubsection{Cosine similarity}
Cosine similarity measures the angle between the two vectors whose coordinates are given by the scores in the evaluated and ground truth rankings.
What is evaluated here is the alignment between rankings, i.e., the extent to which the candidate method assigns high/low scores to the same words that receive high/low scores in the ground truth.
Since all rankings have positive scores, cosine similarity is in $[0,1]$, where 0 indicates that the two vectors are orthogonal and 1 means that they are perfectly aligned.
Since we are computing the value of an angle, only the ratios/proportions of scores matter here.
E.g., the two rankings $X=[1,2,3]$ and $Y=[10,20,30]$ would be considered perfectly aligned.

\subsubsection{Spearman's rho}
Spearman's rho \cite{spearman1904proof} is a measure of rank correlation.
More precisely, it equals the famous Pearson product-moment correlation coefficient ($r=\nicefrac{\text{cov}(X,Y)}{\sigma_X \sigma_Y}$) computed from the \textit{ranks} of the scores in the two rankings, rather than on the scores themselves.

\subsubsection{Kendall's tau}
Kendall's tau \cite{kendall1938new} is another measure of rank correlation, based on \textit{signs} of ranks.
One can compute it by counting concordant and discordant pairs among the ranks of the scores in the two rankings.
More precisely, given two rankings $X$ and $Y$, a pair $\big((x_i,y_i),(x_j,y_j)\big)$ for $i<j$ is said to be concordant if $x_i<x_j$ and $y_i<y_j$. Based on this notion, the metric is expressed:

\begin{equation}
    \tau = \frac{\text{\#concordant}-\text{\#discordant}}{\binom{n}{2}}
\end{equation}

\noindent Kendall's tau can also be written:

\begin{equation}
    \tau = \binom{n}{2} \sum_{i<j} \text{sign}(x_i - x_j) \text{sign}(y_i - y_j)
\end{equation}

\noindent where {\small \texttt{sign}} designates the sign function and $n$ is the length of the two rankings.

Both Spearman's rho and Kendall's tau take values in $[-1,1]$ (for reversed and same rankings), and approach zero when the correlation between the two rankings is low (independence).

\subsection{Information retrieval metrics}

\subsubsection{p@k}
Here, we simply compute the percentage of words in the top 10\% of the candidate ranking that are present in the top 10\% of the ground truth ranking.
The idea here is to measure ranking quality for the most polysemous words.

\subsubsection{NDCG}
The Normalized Discounted Cumulative Gain or NDCG \cite{jarvelin2002cumulated} is a standard metric in information retrieval.
It is based on the Discounted Cumulative Gain (DCG):

\begin{equation}
\mathrm{DCG} = \sum_{i=1}^{n} \frac{2^{\text{score}_i} - 1}{\mathrm{log}_2(i+1)}
\end{equation}

\noindent where $\text{score}_i$ designates the ground truth score of the word at the $i^{th}$ position in the ranking under consideration, and $n$ denotes the length of the ranking.
NDCG is then expressed as:

\begin{equation}
    \text{NDCG} = \frac{\text{DCG}(\text{candidate})}{\text{DCG}(\text{ground truth})}
\end{equation}

\noindent the denominator is called the \textit{ideal} DCG, or IDCG. It is the DCG computed with the order provided by the ground truth ranking, that is, for the best possible word positioning.

Since the scores are penalized proportionally to their position in the ranking (with some concavity), the more words with high ground truth scores are placed on top of a candidate ranking, the better the NDCG of that ranking.
NDCG is maximal and equal to 1 if the candidate and ground truth rankings are identical.

\subsubsection{RBO}
The Rank Biased Overlap or RBO \cite{webber2010similarity} takes values in $[0,1]$, where 0 means that the two rankings are independent and 1 that they match exactly.
It is computed as:

\begin{equation}
    \text{RBO} = (1-p) \sum_{d=1}^{n} p^{d-1} A_d
\end{equation}

\noindent where $A_d$ is the proportion of words belonging to both rankings up to position $d$, $n$ is the length of the rankings, and $p$ is a parameter controlling how steep the decline in weights is: the smaller $p$, the more top-weighted the metric is.
When $p=0$, only the top-ranked word is considered, and the RBO is either zero or one. When $p$ is close to 1, the weights become flat, and more and more words are considered.
We used $p=0.98$ in our experiments, which means that the top 50 positions received 86\% of all weight.

\section{Implementation details}
To compute our scores, we built on the code of the pyramid match kernel from the \texttt{GraKeL} Python library \citep{siglidis2018grakel}. 
We used the base R \cite{baseR} \texttt{cor()} function\footnote{\url{https://stat.ethz.ch/R-manual/R-patched/library/stats/html/cor.html}} to compute the $\tau$ and $\rho$ statistics.
For RBO, we relied on a publicly available Python implementation\footnote{ \url{https://github.com/changyaochen/rbo}}.
For all other metrics, we wrote our own implementations.
Full details about design choices, tokenizers, stopword lists etc., can also be found in our publicly available code repository: \url{https://github.com/ksipos/polysemy-assessment}.

\begin{figure*}
\centering
\subfloat[]{\includegraphics[width=.49\textwidth]{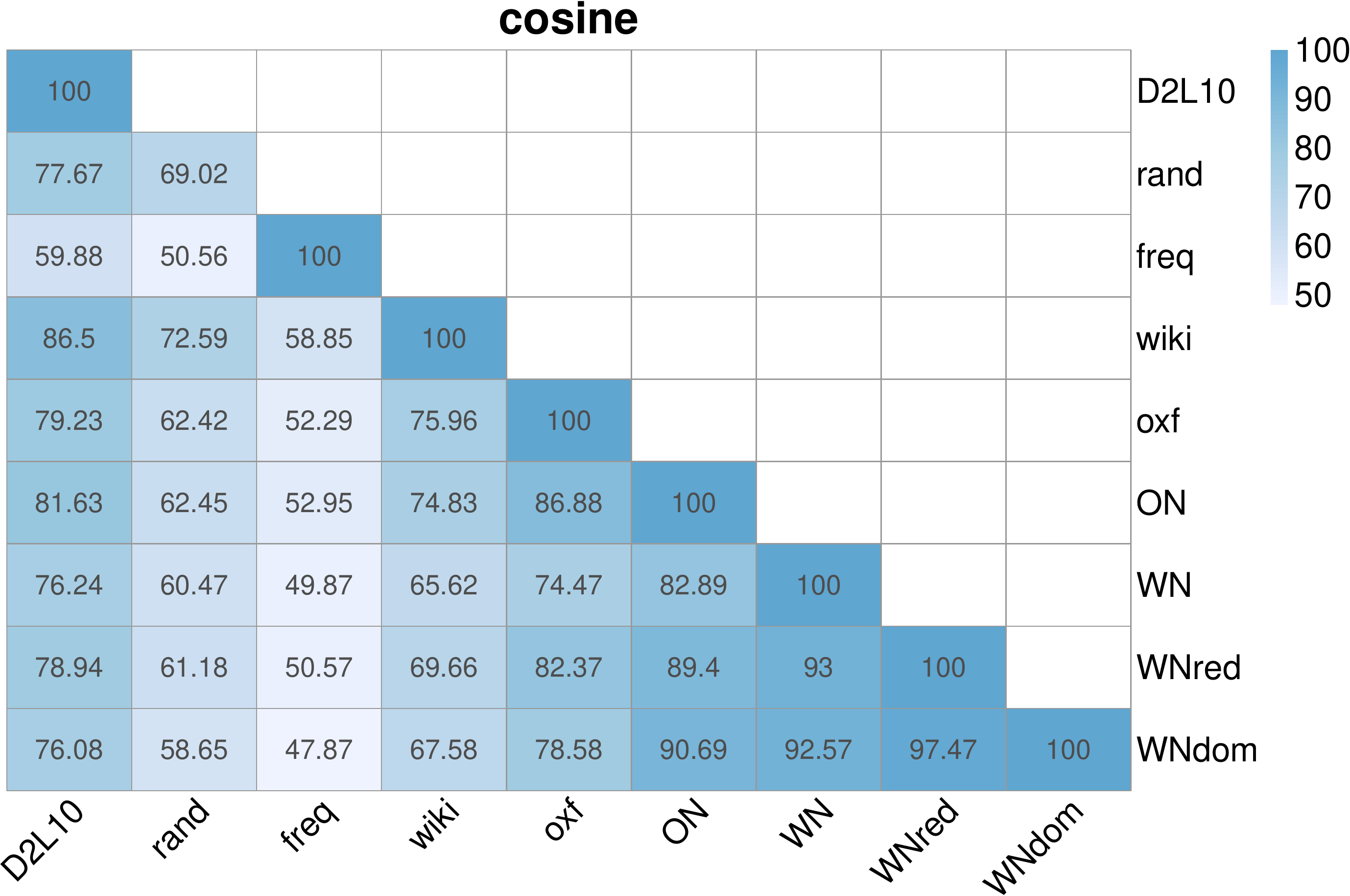}}\hfill
\subfloat[]{\includegraphics[width=.49\textwidth]{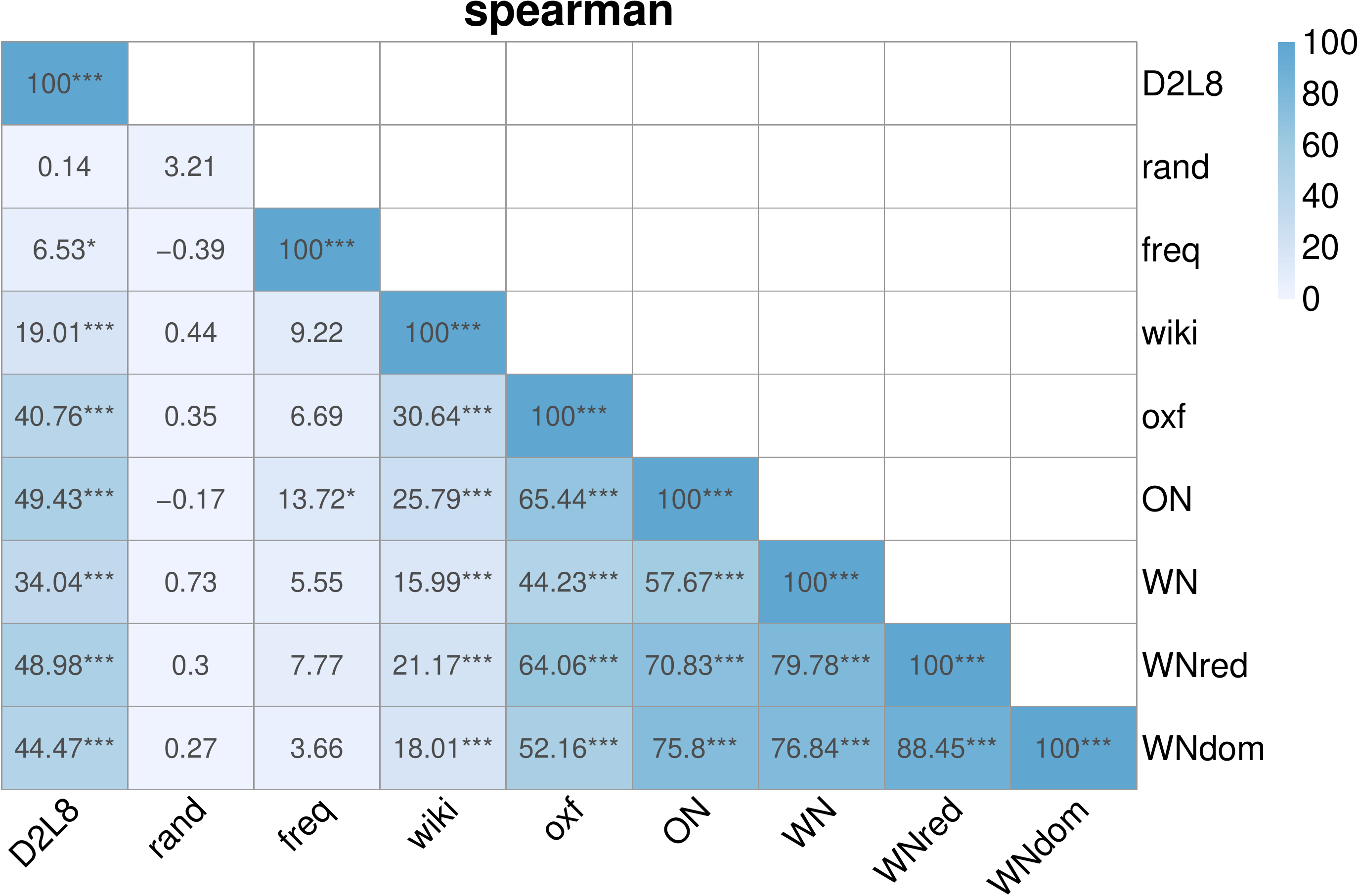}}\hfill \\
\subfloat[]{\includegraphics[width=.49\textwidth]{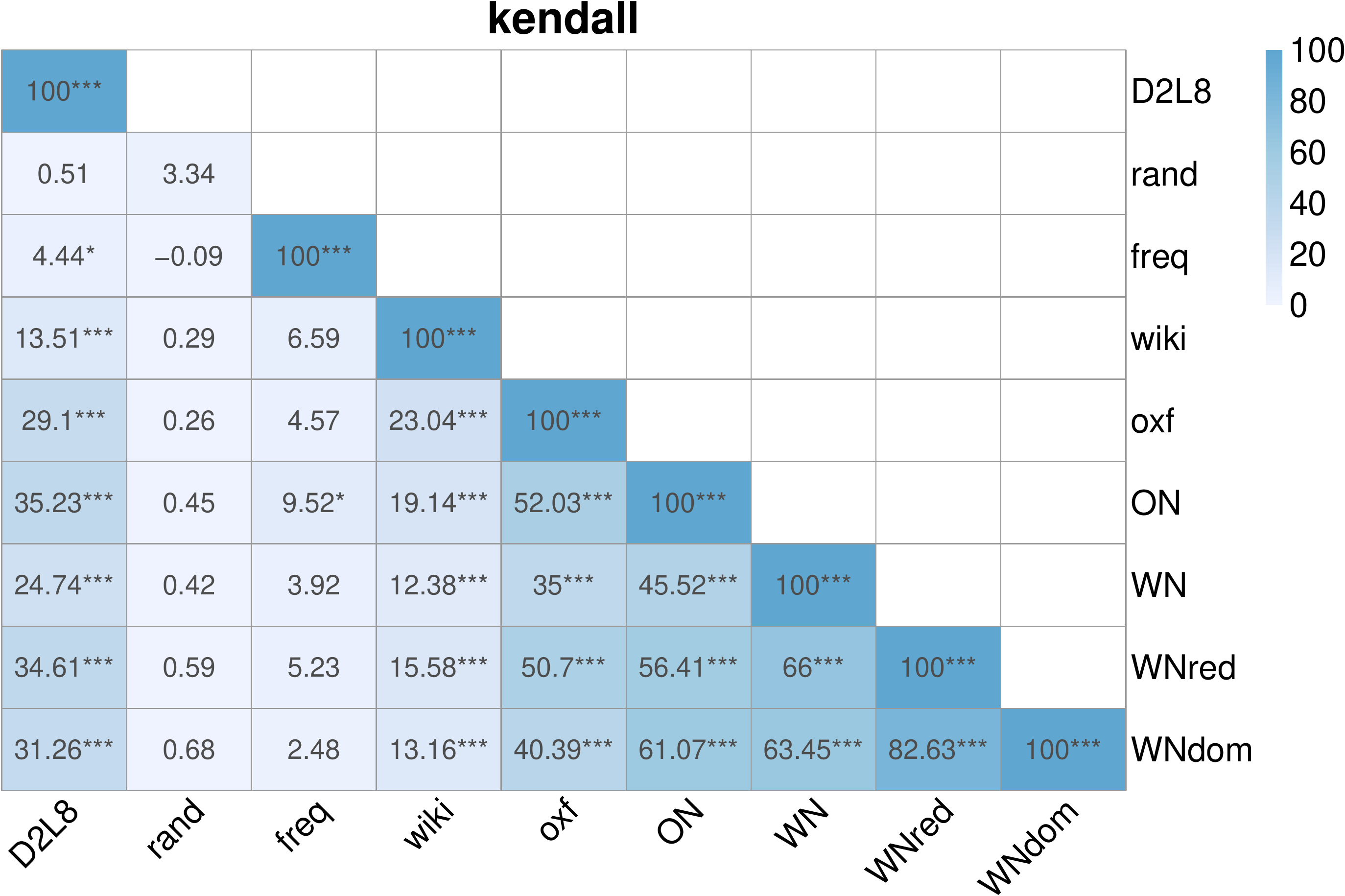}}
\subfloat[]{\includegraphics[width=.49\textwidth]{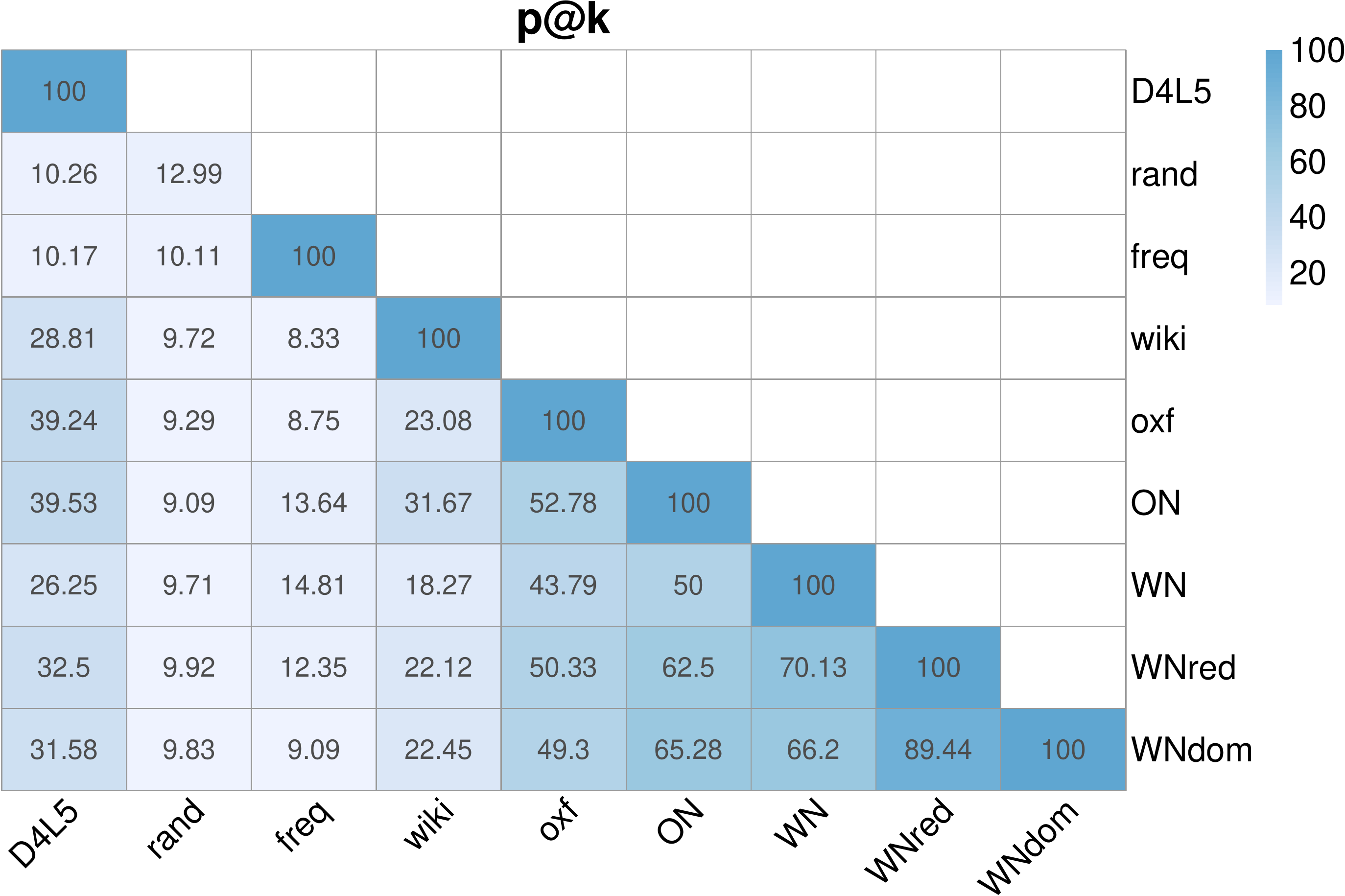}}\hfill \\
\subfloat[]{\includegraphics[width=.49\textwidth]{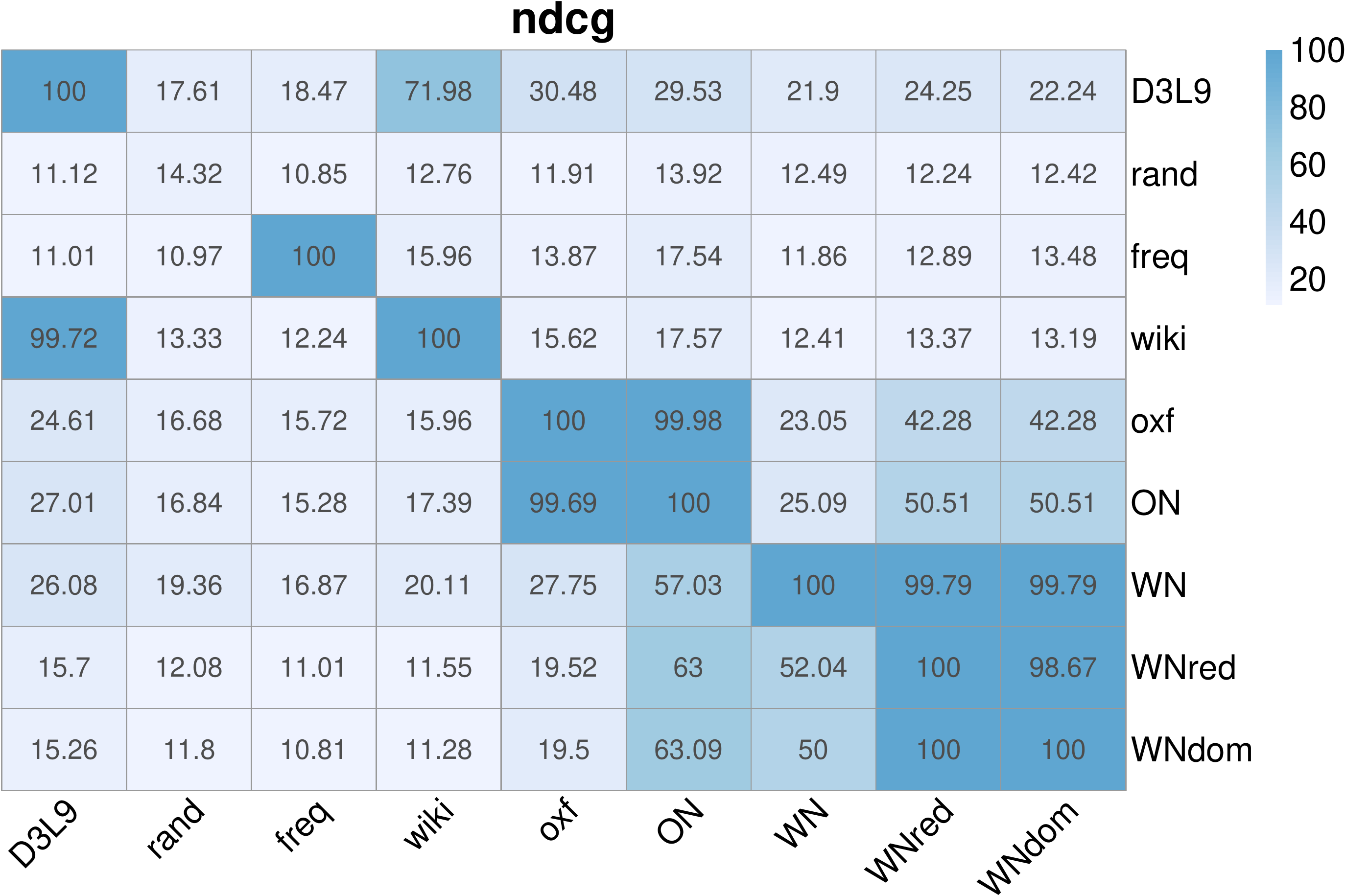}}\hfill
\subfloat[]{\includegraphics[width=.49\textwidth]{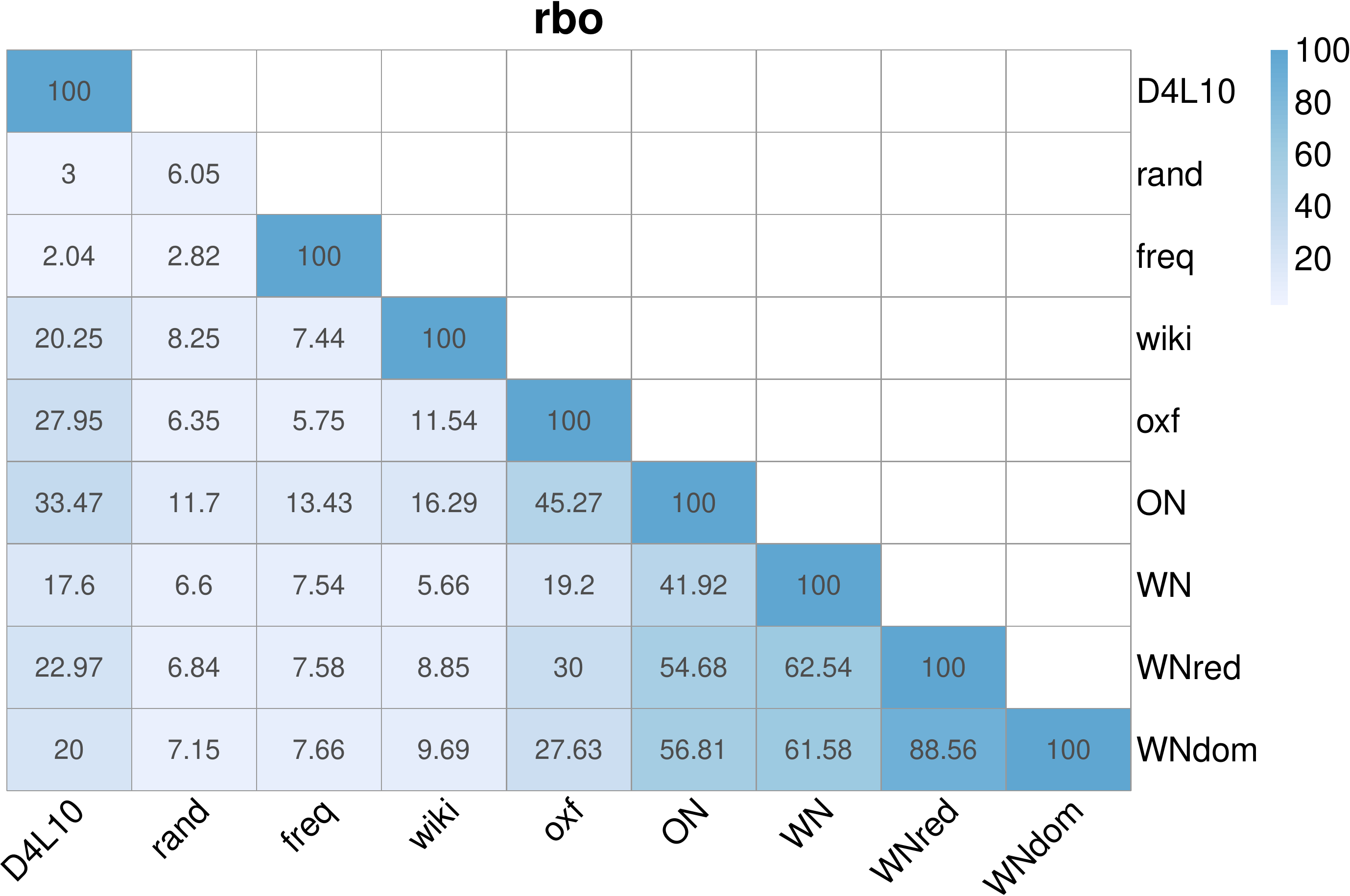}}
\captionsetup{size=small}
\caption{Pairwise similarity matrices between methods.
For readability, all scores are shown as percentages.
For Kendal and Spearman, * and *** mean statistical significance at $p\leq0.01$ and $p\leq0.0001$, respectively.
For a given metric, our configuration that best matches (on average) all other methods (except random and frequency) is always shown first.
$D$D$L$L means that the compressed contextual embedding space has D dimensions and that the hierarchy has L levels.
Rand, freq, wiki, oxf, ON, WN, WNred, and WNdom are short for random, frequency, Wikipedia, Oxford, OntoNotes, WordNet, WordNet reduced, and WordNet domains.
All metrics except NDCG are symmetric, hence we only show one triangle for them.
For NDCG, candidate methods are shown as columns and ground truths as rows.}
\label{fig:results}
\end{figure*}

\begin{table*}
\small
\centering
\begin{tabular}{c|c}
\hline
Sentences & Bin coordinates \\
\hline
it stars christopher lee as \textbf{count} dracula along with dennis waterman & $(3,5,1)$ \\
the \textbf{count} of the new group is the sum of the separate counts of the two original groups & $(4,1,3)$ \\
the first fight did not \textbf{count} towards the official record & $(4,5,1)$ \\
\hline
five year old horatia came to \textbf{live} at merton in may 1805 & $(2,5,2)$ \\
it features various amounts of \textbf{live} and backstage footage while touring & $(4,2,4)$ \\
\hline
first tax bills were used to pay taxes and to register \textbf{bank} deposits and bank credits & $(4,2,4)$ \\
the ball nest is built on a \textbf{bank} tree stump or cavity & $(5,2,3)$
\end{tabular}
\captionsetup{size=small}
\caption{Sentences containing different senses of the same word can be sampled by selecting from different bins. \label{table:sents}}
\end{table*}

\begin{table*}
\small
\centering
\begin{tabular}{c|c}
\hline
Keywords & Bin coordinates \\
\hline
also, gas, used, system, protein, blood, new, steel, food, made & $(20,11,16,9)$ \\
\hline
first, new, one, second, later, world, national, olympic, team, games & $(19,13,15,13)$ \\
\hline
album music rock one labour number chart songs metal single & $(21,14,14,16)$
\end{tabular}
\captionsetup{size=small}
\caption{Towards automatic word sense induction: top 10 most frequent words for different bins containing the word \textit{metal}. The bins correspond to 3 senses of metal: chemical element, Olympics (medals), and music. \label{table:wsi}}
\end{table*}

\section{Results and observations}\label{sec:results}

\noindent \textbf{Our rankings correlate well with human rankings}.
Results are shown in Fig. \ref{fig:results}, as pairwise similarity matrices, for all six metrics.
For readability, all scores are shown as percentages.
For a given metric, our configuration that best matches, on average, all other methods (except random and frequency) is always shown as the first column.
Since all metrics except NDCG are symmetric, we only show the lower triangles of the other matrices.
For NDCG, candidate methods are shown as columns and ground truths as rows.

For each of the six evaluation metrics, it can be seen that the ranking generated by our unsupervised, data-driven method is well correlated with all human-derived ground truth rankings.
This means that our method is robust to how one defines and measures correlation or similarity.

In some cases, we even very closely reproduce the human rankings.
For instance, our best configurations for cosine and NDCG get almost perfect scores of 86.5 and 99.72 when compared against Wikipedia.
In terms of Kendall's tau, Spearman's rho, p@k, and RBO, we are also very close to OntoNotes (scores of 49.43, 35.23, 39.53, and 33.47, resp.).

Finally, the correlation between our rankings and the human rankings can also be observed to be, everywhere, much stronger than that between the baseline rankings (random and frequency) and the human rankings.

\noindent \textbf{Statistical significance}.
We computed statistical significance for the Spearman's rho and Kendall's tau metrics.
As shown in Fig. \ref{fig:results}, the null hypothesis that there is no correlation between our rankings and the human-derived ground truth rankings was systematically rejected everywhere, with very high significance ($p \leq 0.0001$).

However, against the random baseline, the same null hypothesis (no correlation) was accepted everywhere.
Against frequency, the null was rejected, but very weakly (only at the $p \leq 0.01$ level), and with very low correlation coefficients (6.53 for Spearman and 4.44 for Kendall).

Finally, the correlation between the random and frequency rankings and the ground truth rankings is never statistically significant, except for the pair frequency/OntoNotes, but again, at a weak level ($p \leq 0.01$).

\begin{figure*}
\centering
\subfloat[]{\includegraphics[width=.25\textwidth]{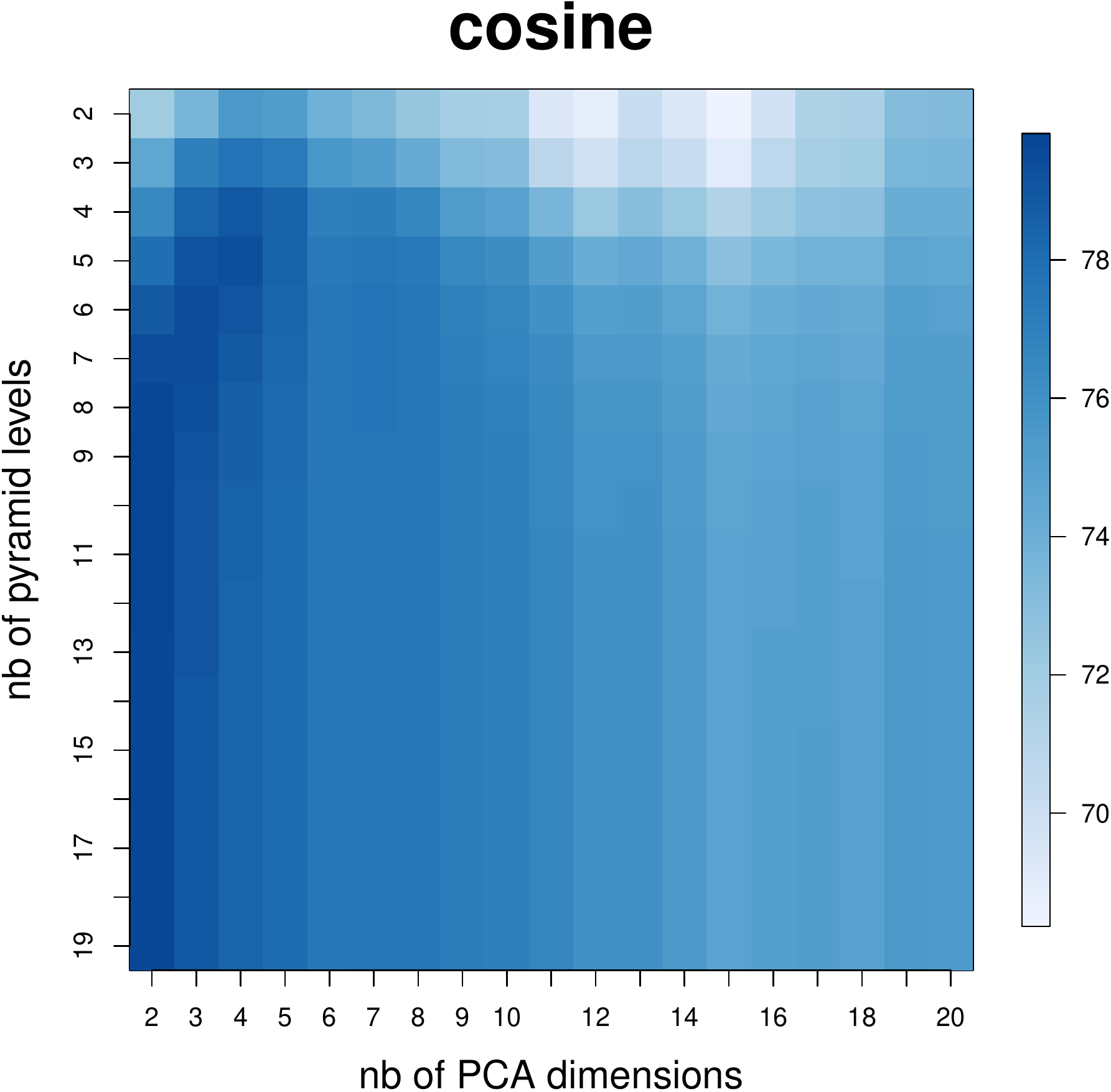}}
\subfloat[]{\includegraphics[width=.25\textwidth]{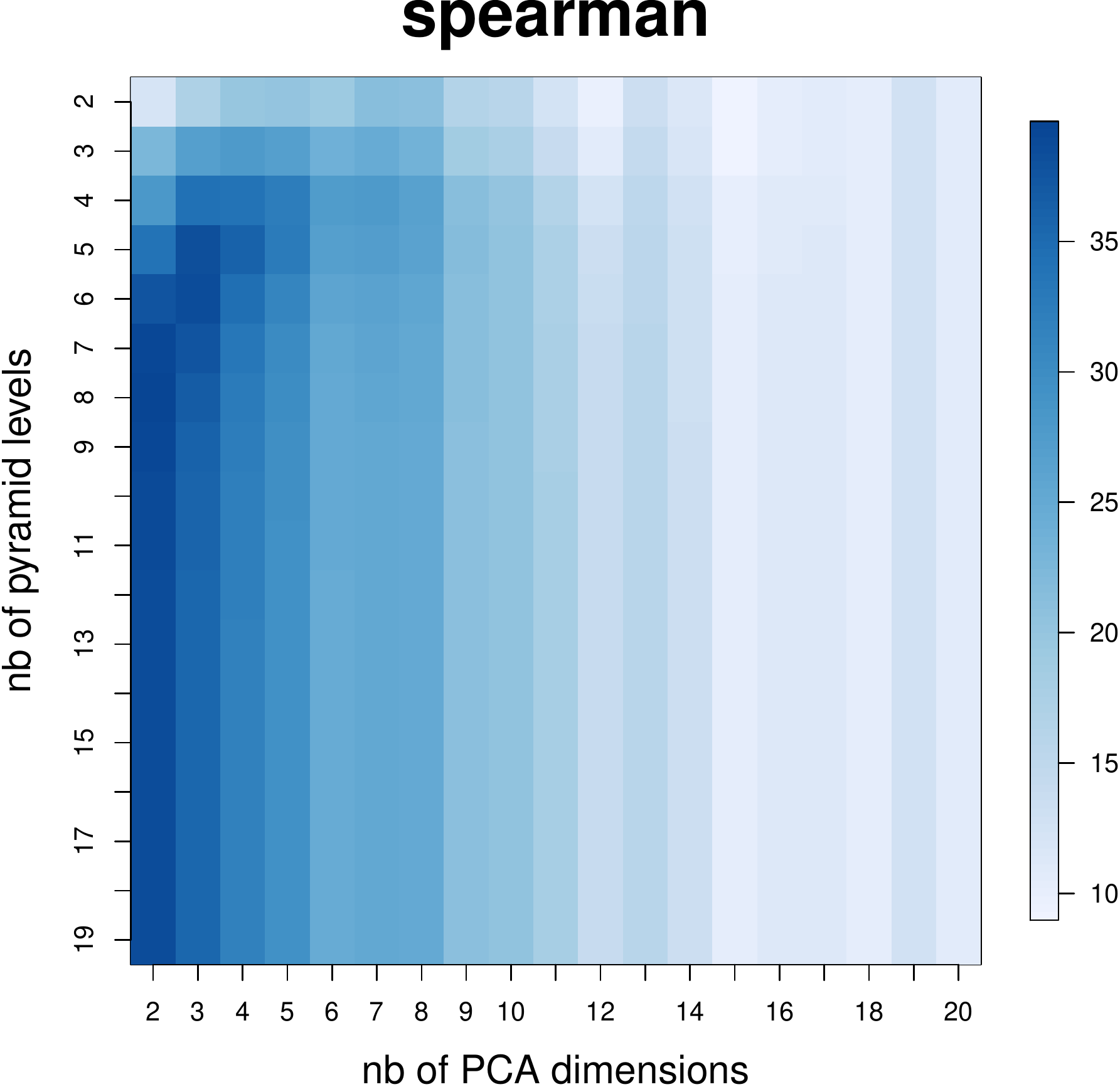}}
\subfloat[]{\includegraphics[width=.25\textwidth]{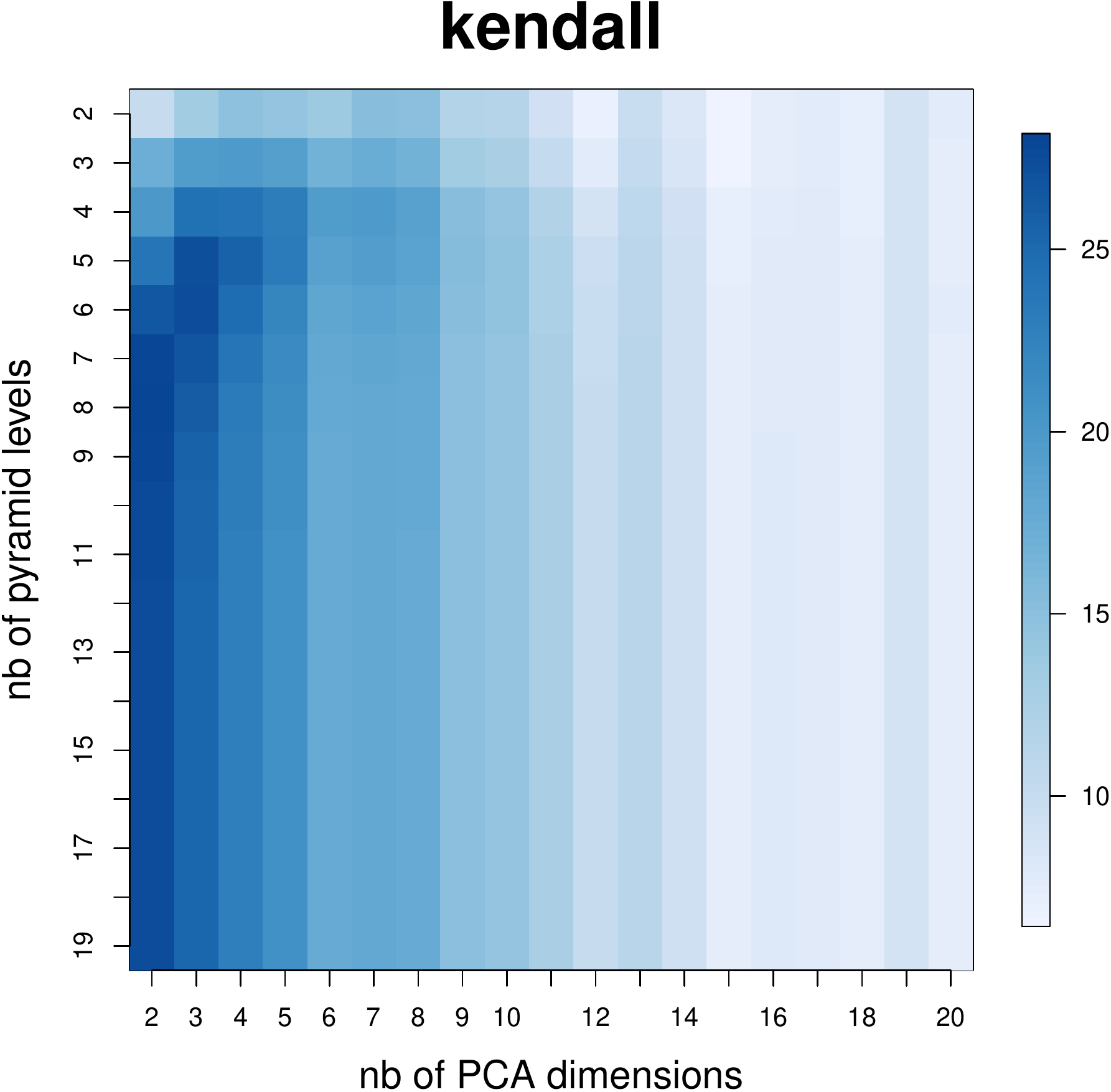}} \\
\subfloat[]{\includegraphics[width=.25\textwidth]{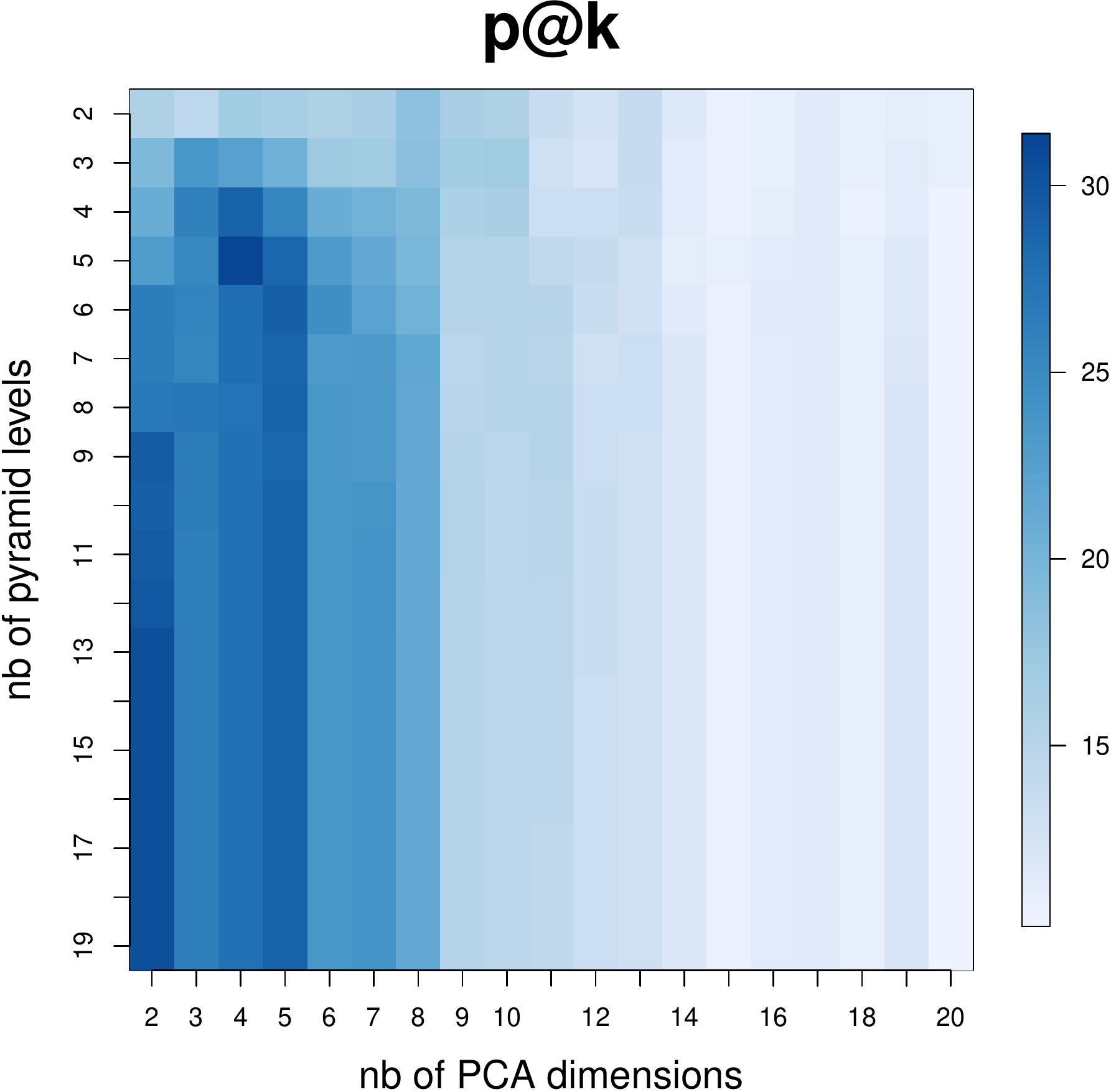}}
\subfloat[]{\includegraphics[width=.25\textwidth]{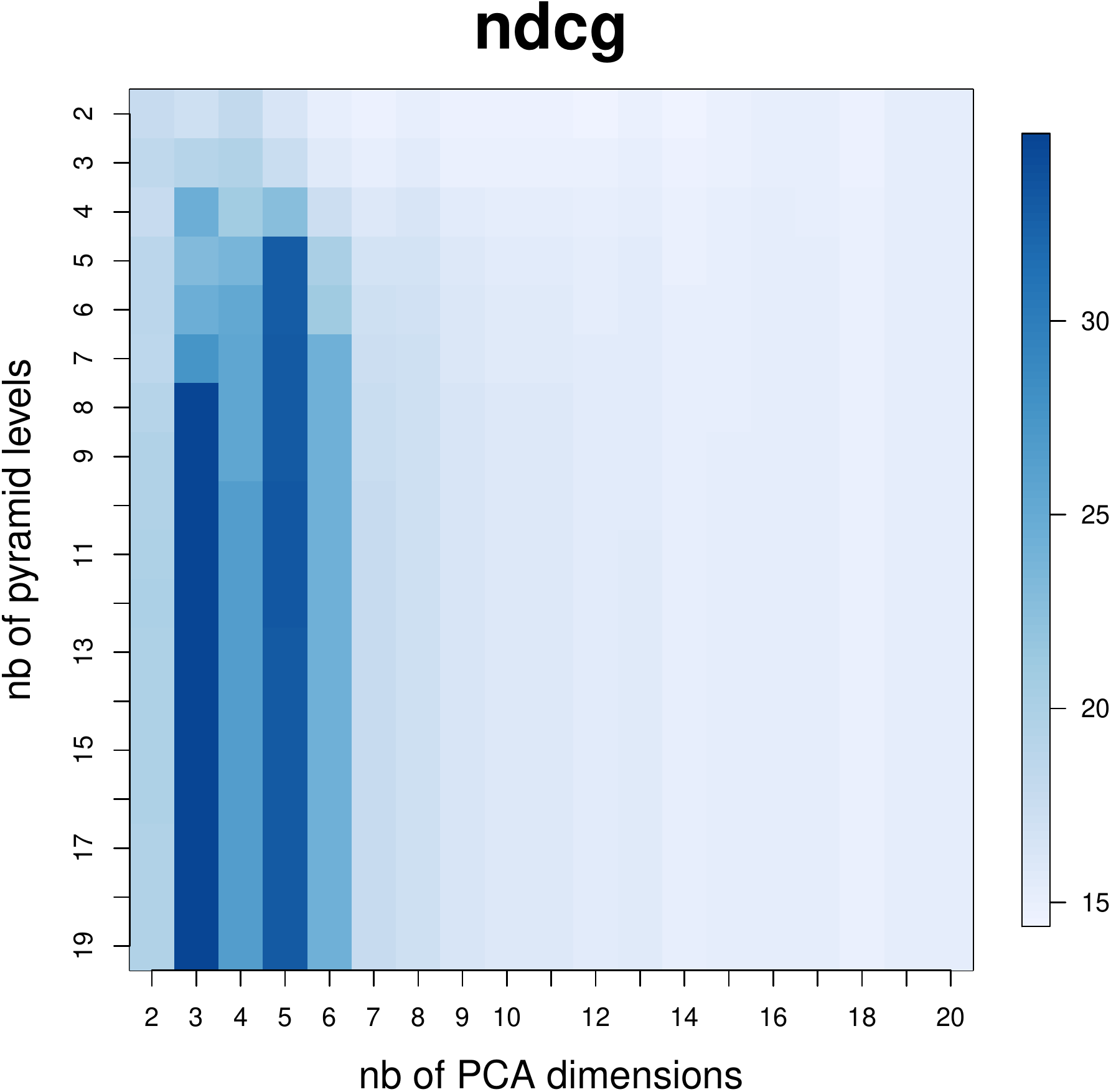}}
\subfloat[]{\includegraphics[width=.25\textwidth]{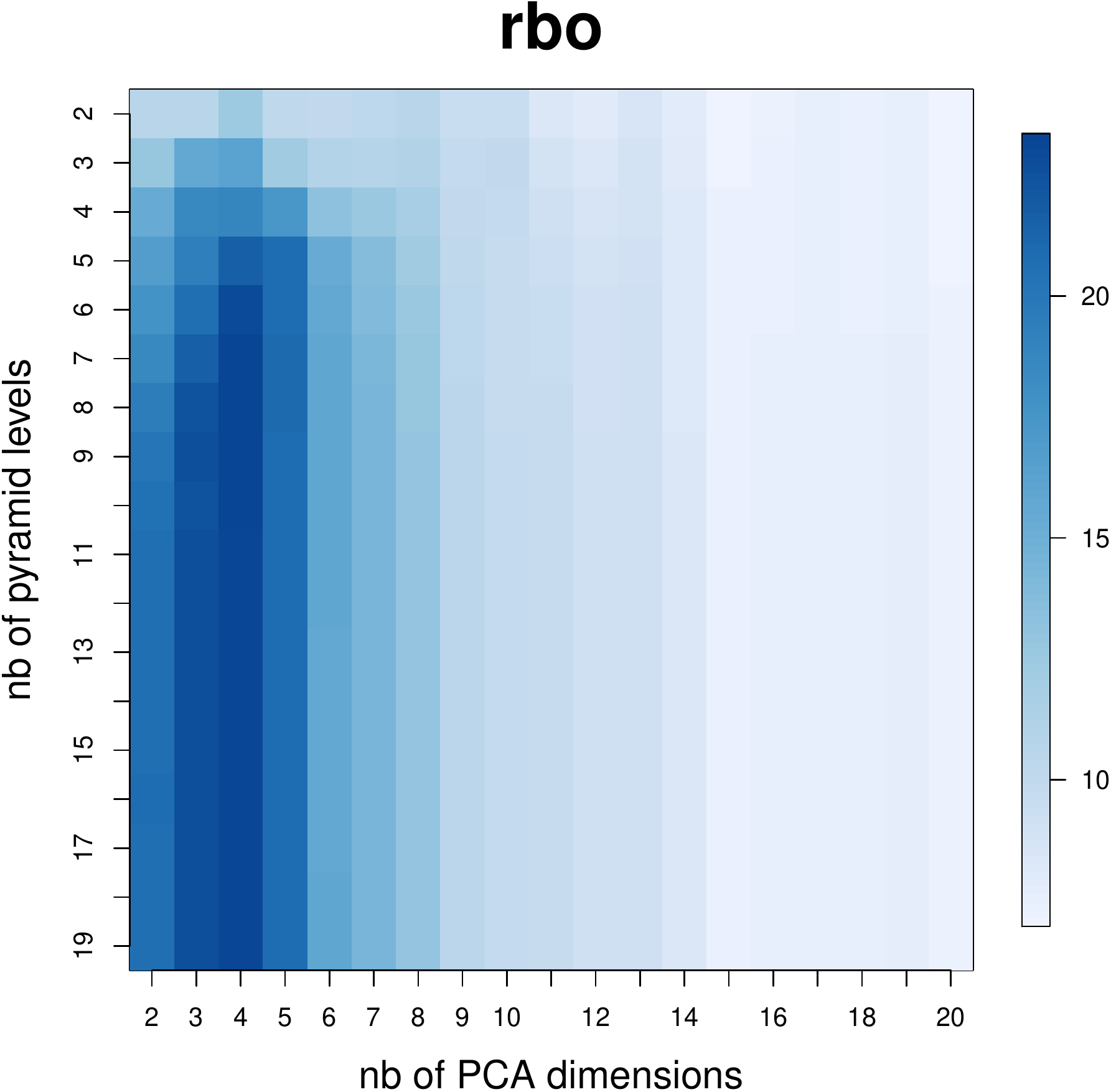}}
\captionsetup{size=small}
\caption{Performance (color scale) vs. number of PCA dimensions ($x$ axis) vs. number of levels in the hierarchy ($y$ axis).}
\label{fig:grid_search}
\end{figure*}

\begin{figure*}
\centering
\subfloat[]{\includegraphics[width=.145\textwidth]{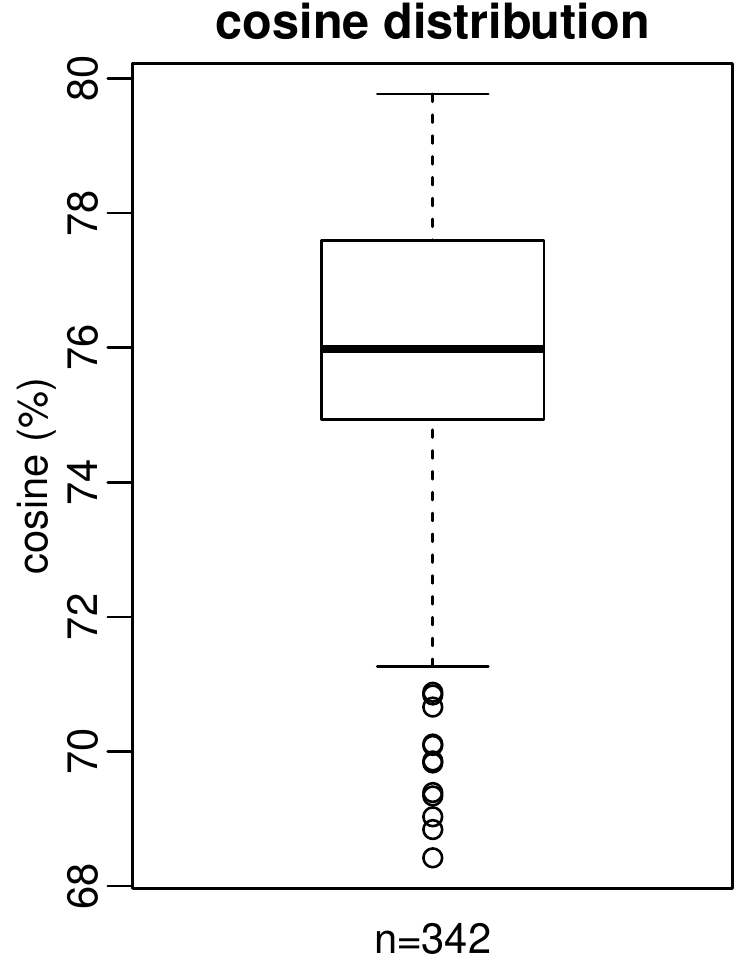}}
\subfloat[]{\includegraphics[width=.145\textwidth]{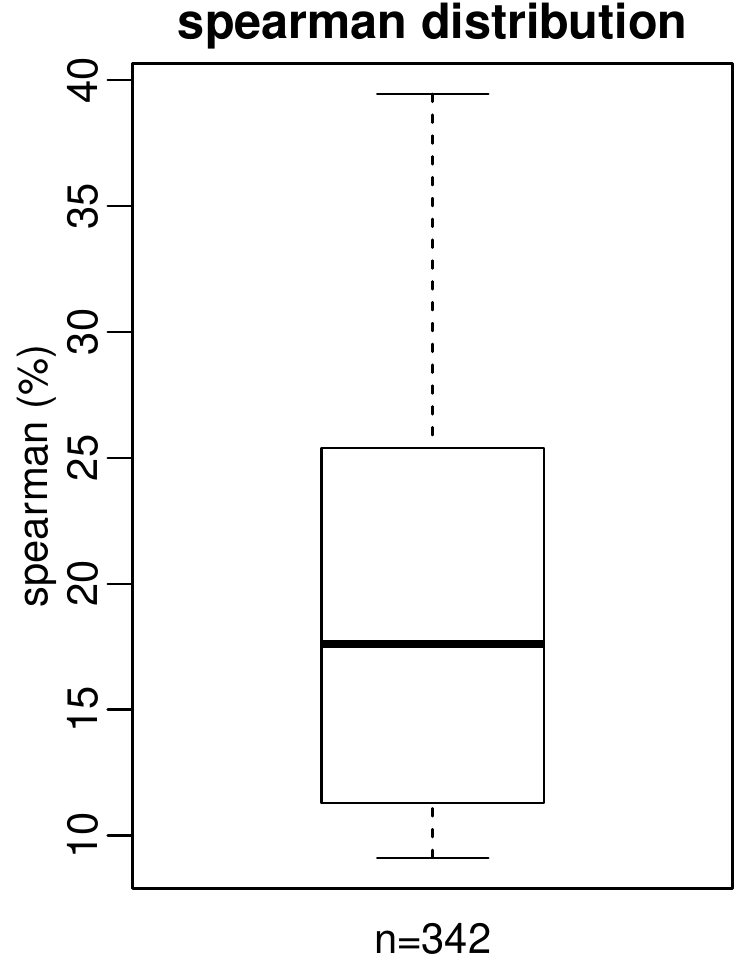}}
\subfloat[]{\includegraphics[width=.145\textwidth]{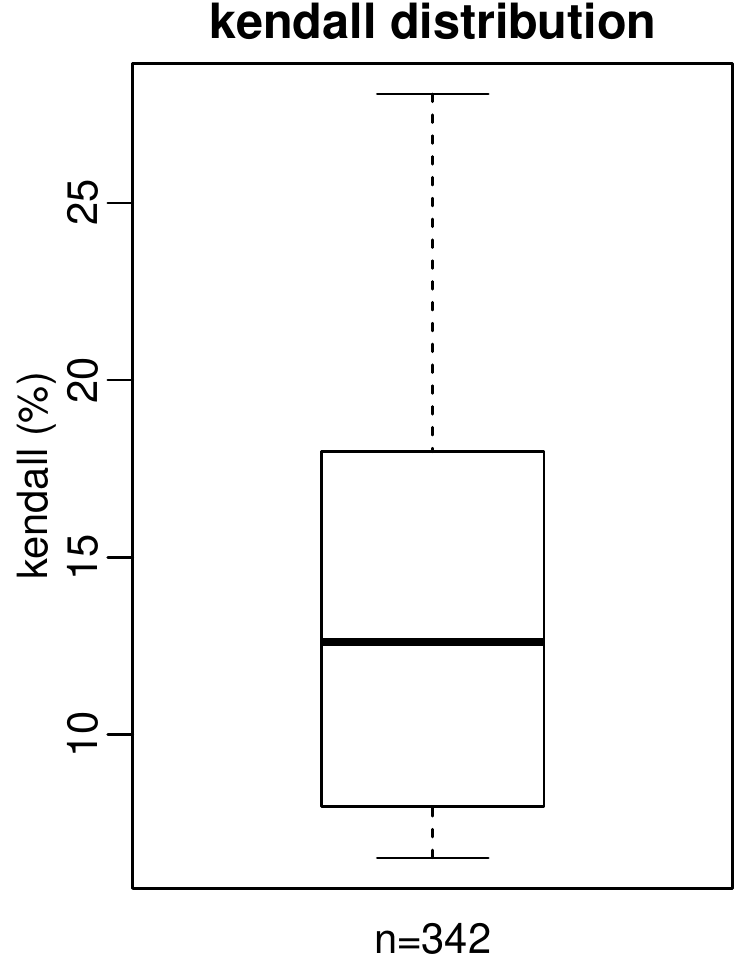}}
\subfloat[]{\includegraphics[width=.145\textwidth]{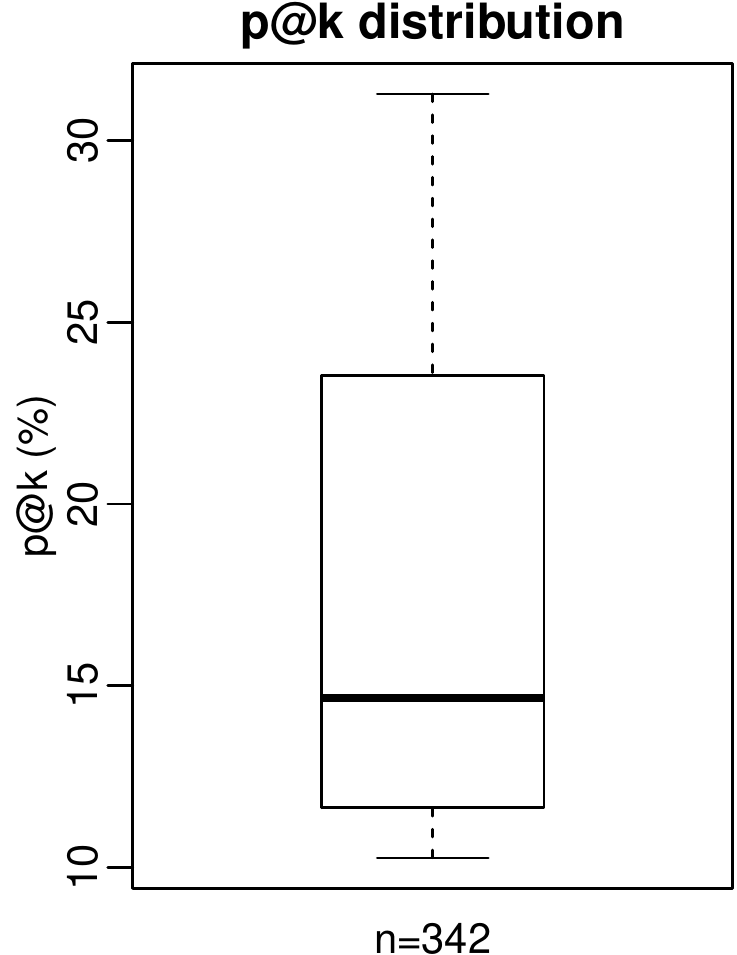}}
\subfloat[]{\includegraphics[width=.145\textwidth]{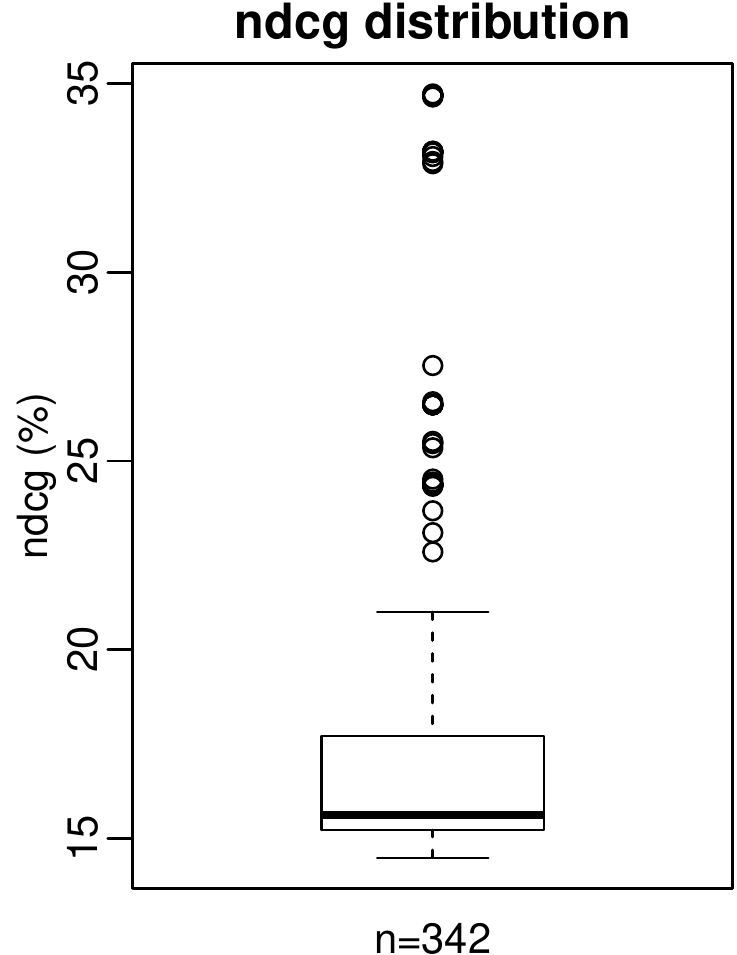}}
\subfloat[]{\includegraphics[width=.145\textwidth]{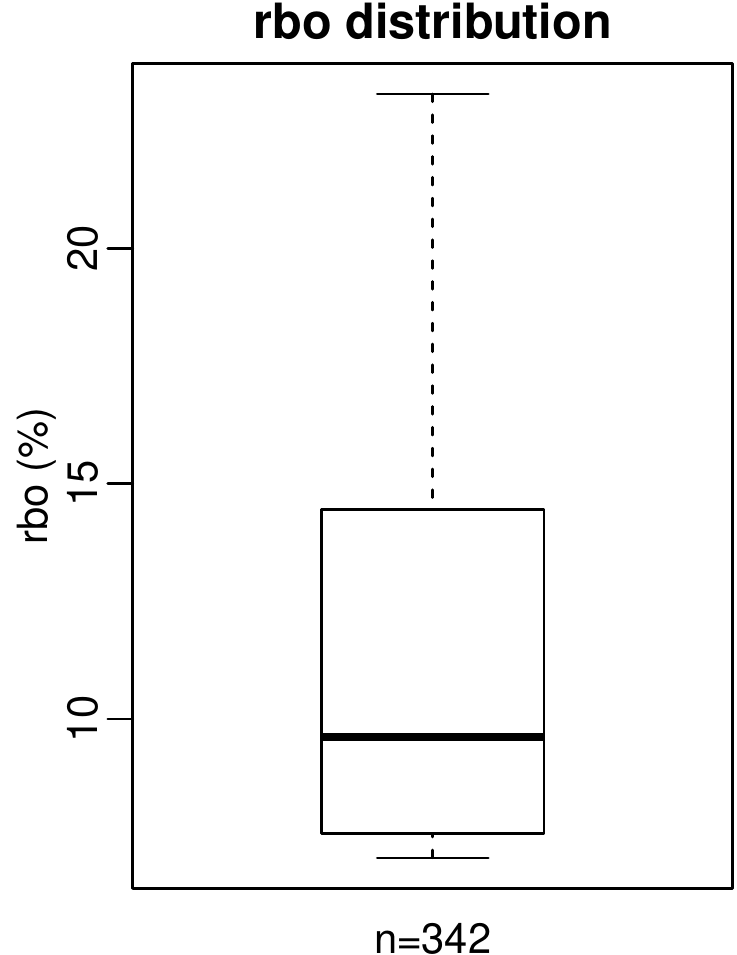}}
\captionsetup{size=small}
\caption{Performance distributions over the 342 values in the discrete hyperparameter space (grids of Fig. \ref{fig:grid_search}).}
\label{fig:boxplots}
\end{figure*}

\noindent \textbf{Hyperparameters have a significant impact on performance, but optimal values are consistent across metrics}.
First, as can be observed from Fig. \ref{fig:grid_search} and Fig. \ref{fig:boxplots}, there is a large variability in performance when $D$ (number of PCA dimensions) and $L$ (number of levels in the hierarchy) vary.

However, for all six evaluation metrics, the best configurations are very similar: $D2L10$, $D2L8$, $D2L8$, $D4L5$, $D3L9$, and $D4L10$\footnote{for RBO, $D4L10$ and $D4L8$ had the same score.}.
Given the rather large grid we explored ($[2,20] \times [2,19]$ for $D$ and $L$, resp.), with 342 combinations in total, we can say that all these optimal values belong to the same small neighborhood.
This interpretation is confirmed by inspecting Fig. \ref{fig:grid_search}, where it can clearly be seen that the optimal area of the hyperparameter space is robust to metric selection and consistently corresponds to small values of $D$ (around 3), and values of $L$ at least above 3 or 4, ideally around 8.
For larger values of $L$, performance plateaus (keeping $D$ fixed).
In other words, it is necessary to have some levels in the hierarchy, but having very deep hierarchies is not required for our method to work well.
A benefit of having such small optimal values of $D$ and $L$ is their affordability, from a computational standpoint.

\noindent \textbf{All rankings derived from WordNet-based resources are highly correlated}.
It is interesting to note that the rankings generated from OntoNotes, WordNet, WordNet reduced, and WordNet domains, all are highly similar.
And this, despite the very different sense granularities they have.
This means that despite the apparent differences in these resources, they all tend to produce similar polysemy rankings.
The Oxford rankings tend to be part of this high-similarity cluster as well, to a lesser extent.

\noindent \textbf{Frequent words are not the most polysemous}.
Finally, one last interesting observation is that while the frequency ranking is much better than the random ones, it still is far away from the human rankings.
In other words, the frequency of appearance of a word (excluding stopwords, of course) is not as good an indicator of its polysemy as one could expect.
Some words that follow this observation are "number", "population", and "war".

\noindent \textbf{A note on ties}. To assess the impact of ties on the reported results, we repeated all of our experiments multiple times with different tie-breaking strategies (e.g., random, alphabetical...). Results do not change: we find the same best parameter combinations, and the differences in the similarity matrices are minimal.

\section{Other applications}\label{sec:app}
\noindent \textbf{Sampling diverse examples}. An interesting by-product of our discretization strategy is that it can be used to select sentences containing different senses of the same word, as illustrated in Table \ref{table:sents}.
Provided a mapping, for a given word, between the sentences that were passed to the pre-trained language model and the vectors, we can sample vectors from different bins and retrieve the associated sentences.
If the bins are distant enough, the sentences will contain different senses of the word.
For instance, in Table \ref{table:sents}, we can see that we are able to sample sentences containing three senses of the word \textit{count}: (1) noble title, (2) determining the total number of, and (3) taking into account.
This has many useful applications in practice, e.g., in information retrieval, NLG and conversational systems, dataset creation, etc.

\noindent \textbf{Automatic word sense induction}.
A simple way of capitalizing on our binning strategy to create word sense inventories would consist in (1) selecting distant bins for a given word, and (2) labeling the selected bins with senses. Both steps can be performed automatically.
While this will be investigated in future work, we still give, as a proof of concept, an example in Table \ref{table:wsi}. In this example, keywords are extracted from distant bins containing the word \textit{metal}, and different senses are retrieved.

\section{Related work}\label{sec:related}

\noindent \textbf{Task}.
Several previous efforts have interested themselves in creating sense inventories without human experts.
As an example, in \citet{rumshisky2011crowdsourcing, rumshisky2012word}\footnote{\label{foot:email}We asked the authors to share annotations with us to use as ground truth, but they were unable to do so.},
Amazon Mechanical Turk (AMT) workers are given a set of sentences containing the target word and one sentence that is randomly selected from this set as a target sentence.
Workers are then asked to judge, for each sentence, whether the target word is used in the same way as in the target sentence.
This creates an undirected graph of sentences where clustering can be applied to find senses.
To label clusters with senses, one has to inspect the sentences in each cluster manually.

More recently, \citet{jurgens2013embracing}\footnote{same as footnote \ref{foot:email}.} compared three annotation methodologies for gathering word sense labels on AMT.
The methods compared are Likert scales, two-stage select and rate, and the difference between counts of when senses were rated best/worst.
Regardless of the strategy, inter-annotator agreement remains low (around 0.3).

\noindent \textbf{Methodology}.
In the original ELMo paper, \citet{peters2018deep} have shown that using contextual word representations (through nearest-neighbor matching) improves word sense disambiguation.
\citet{hadiwinoto2019improved,coenen2019visualizing} showed that this technique works well with BERT too.
\citet{pasini2020clubert} uses a combination of BERT embeddings and a knowledge-based WSD model to generate word sense distributions, while \citet{giulianelli-etal-2020-analysing} uses clustering over the embeddings to detect semantic shifts.

Our approach is also related in spirit to pyramid matching \citep{nikolentzos2017matching, grauman2007pyramid, lazebnik2006beyond}.
This kernel-based method originated in computer vision. It computes the similarity between objects by placing a sequence of increasingly coarser grids over the feature space and taking a weighted sum of the number of matches occurring at each level.
Matches found at finer resolutions are weighted more than matches found at coarser resolutions.

\section{Conclusion}
We proposed a novel unsupervised, fully data-driven geometrical approach to estimate word polysemy.
Our approach builds multiresolution grids in the contextual embedding space.
Through rigorous experiments, we showed that our rankings are well correlated (with strong statistical significance) to 6 different human rankings, for 6 different metrics.
Such fully data-driven rankings of words according to polysemy can help in creating new sense inventories, but also in validating and interpreting existing ones.
Increasing the quality and consistency of sense inventories is a key first step of the word sense disambiguation pipeline.
We also showed that our discretization could be used, at no extra cost, to sample contexts containing different senses of a given word, which has useful applications in practice.
Finally, the unsupervised nature of our method makes it applicable to any language.

While our scores are a good proxy for polysemy, they are not equal to word sense counts.
Moreover, we do not label each sense.
Future work should address these challenges by, e.g., automatically selecting bins of interest and generating labels for them (see section \ref{sec:app}).

Future work should also perform some sort of extrinsic evaluation.
For instance, the Word-in-Context task \cite{pilehvar2018wic} could be used, where two occurrences would be classified as having the same meaning if their two vectors fall into the same bin.

Another direction is investigating how different contextual embeddings (e.g., BERT, BART) impact our rankings, including in languages other than English \cite{eddine2020barthez, CaneteCFP2020}, and low-resource languages.

\section{Acknowledgments}
We thank the three anonymous reviewers for their helpful feedback, and Giannis Nikolentzos for helpful discussions about pyramid matching.
The GPU used in this study was donated by the Nvidia corporation as part of their GPU grant program.
This work was supported by the \href{https://linto.ai/}{LinTo} project.

\bibliography{emnlp2018}

\begin{thebibliography}{35}
\expandafter\ifx\csname natexlab\endcsname\relax\def\natexlab#1{#1}\fi

\bibitem[{Artstein and Poesio(2008)}]{artstein2008inter}
Ron Artstein and Massimo Poesio. 2008.
\newblock Inter-coder agreement for computational linguistics.
\newblock \emph{Computational Linguistics}, 34(4):555--596.

\bibitem[{Bentivogli et~al.(2004)Bentivogli, Forner, Magnini, and
  Pianta}]{bentivogli2004revising}
Luisa Bentivogli, Pamela Forner, Bernardo Magnini, and Emanuele Pianta. 2004.
\newblock Revising the wordnet domains hierarchy: semantics, coverage and
  balancing.
\newblock In \emph{Proceedings of the workshop on multilingual linguistic
  resources}, pages 94--101.

\bibitem[{Brown et~al.(2010)Brown, Rood, and Palmer}]{brown2010number}
Susan~Windisch Brown, Travis Rood, and Martha Palmer. 2010.
\newblock Number or nuance: Which factors restrict reliable word sense
  annotation?
\newblock In \emph{LREC}.

\bibitem[{Cañete et~al.(2020)Cañete, Chaperon, Fuentes, Ho, Kang, and
  Pérez}]{CaneteCFP2020}
José Cañete, Gabriel Chaperon, Rodrigo Fuentes, Jou-Hui Ho, Hojin Kang, and
  Jorge Pérez. 2020.
\newblock Spanish pre-trained bert model and evaluation data.
\newblock In \emph{PML4DC at ICLR 2020}.

\bibitem[{Coenen et~al.(2019)Coenen, Reif, Yuan, Kim, Pearce, Vi{\'e}gas, and
  Wattenberg}]{coenen2019visualizing}
Andy Coenen, Emily Reif, Ann Yuan, Been Kim, Adam Pearce, Fernanda Vi{\'e}gas,
  and Martin Wattenberg. 2019.
\newblock Visualizing and measuring the geometry of bert.
\newblock \emph{arXiv preprint arXiv:1906.02715}.

\bibitem[{Devlin et~al.(2018)Devlin, Chang, Lee, and
  Toutanova}]{devlin2018bert}
Jacob Devlin, Ming-Wei Chang, Kenton Lee, and Kristina Toutanova. 2018.
\newblock Bert: Pre-training of deep bidirectional transformers for language
  understanding.
\newblock \emph{arXiv preprint arXiv:1810.04805}.

\bibitem[{Eddine et~al.(2020)Eddine, Tixier, and
  Vazirgiannis}]{eddine2020barthez}
Moussa~Kamal Eddine, Antoine J-P Tixier, and Michalis Vazirgiannis. 2020.
\newblock Barthez: a skilled pretrained french sequence-to-sequence model.
\newblock \emph{arXiv preprint arXiv:2010.12321}.

\bibitem[{Ethayarajh(2019)}]{ethayarajh2019contextual}
Kawin Ethayarajh. 2019.
\newblock How contextual are contextualized word representations? comparing the
  geometry of bert, elmo, and gpt-2 embeddings.
\newblock \emph{arXiv preprint arXiv:1909.00512}.

\bibitem[{Giulianelli et~al.(2020)Giulianelli, Del~Tredici, and
  Fern{\'a}ndez}]{giulianelli-etal-2020-analysing}
Mario Giulianelli, Marco Del~Tredici, and Raquel Fern{\'a}ndez. 2020.
\newblock Analysing lexical semantic change with contextualised word
  representations.
\newblock In \emph{Proceedings of the 58th Annual Meeting of the Association
  for Computational Linguistics}, pages 3960--3973, Online. Association for
  Computational Linguistics.

\bibitem[{Grauman and Darrell(2007)}]{grauman2007pyramid}
Kristen Grauman and Trevor Darrell. 2007.
\newblock The pyramid match kernel: Efficient learning with sets of features.
\newblock \emph{Journal of Machine Learning Research}, 8(Apr):725--760.

\bibitem[{Hadiwinoto et~al.(2019)Hadiwinoto, Ng, and
  Gan}]{hadiwinoto2019improved}
Christian Hadiwinoto, Hwee~Tou Ng, and Wee~Chung Gan. 2019.
\newblock Improved word sense disambiguation using pre-trained contextualized
  word representations.
\newblock \emph{arXiv preprint arXiv:1910.00194}.

\bibitem[{Hovy et~al.(2006)Hovy, Marcus, Palmer, Ramshaw, and
  Weischedel}]{hovy2006ontonotes}
Eduard Hovy, Mitch Marcus, Martha Palmer, Lance Ramshaw, and Ralph Weischedel.
  2006.
\newblock Ontonotes: the 90\% solution.
\newblock In \emph{Proceedings of the human language technology conference of
  the NAACL, Companion Volume: Short Papers}, pages 57--60.

\bibitem[{Howard and Ruder(2018)}]{howard2018universal}
Jeremy Howard and Sebastian Ruder. 2018.
\newblock Universal language model fine-tuning for text classification.
\newblock \emph{arXiv preprint arXiv:1801.06146}.

\bibitem[{J{\"a}rvelin and Kek{\"a}l{\"a}inen(2002)}]{jarvelin2002cumulated}
Kalervo J{\"a}rvelin and Jaana Kek{\"a}l{\"a}inen. 2002.
\newblock Cumulated gain-based evaluation of ir techniques.
\newblock \emph{ACM Transactions on Information Systems (TOIS)},
  20(4):422--446.

\bibitem[{Jurgens(2013)}]{jurgens2013embracing}
David Jurgens. 2013.
\newblock Embracing ambiguity: A comparison of annotation methodologies for
  crowdsourcing word sense labels.
\newblock In \emph{Proceedings of the 2013 Conference of the North American
  Chapter of the Association for Computational Linguistics: Human Language
  Technologies}, pages 556--562.

\bibitem[{Kendall(1938)}]{kendall1938new}
Maurice~G Kendall. 1938.
\newblock A new measure of rank correlation.
\newblock \emph{Biometrika}, 30(1/2):81--93.

\bibitem[{Lazebnik et~al.(2006)Lazebnik, Schmid, and
  Ponce}]{lazebnik2006beyond}
Svetlana Lazebnik, Cordelia Schmid, and Jean Ponce. 2006.
\newblock Beyond bags of features: Spatial pyramid matching for recognizing
  natural scene categories.
\newblock In \emph{2006 IEEE Computer Society Conference on Computer Vision and
  Pattern Recognition (CVPR'06)}, volume~2, pages 2169--2178. IEEE.

\bibitem[{Magnini and Cavaglia(2000)}]{magnini2000integrating}
Bernardo Magnini and Gabriela Cavaglia. 2000.
\newblock Integrating subject field codes into wordnet.
\newblock In \emph{LREC}, pages 1413--1418.

\bibitem[{Meng et~al.(2016)Meng, Bradley, Yavuz, Sparks, Venkataraman, Liu,
  Freeman, Tsai, Amde, Owen et~al.}]{meng2016mllib}
Xiangrui Meng, Joseph Bradley, Burak Yavuz, Evan Sparks, Shivaram Venkataraman,
  Davies Liu, Jeremy Freeman, DB~Tsai, Manish Amde, Sean Owen, et~al. 2016.
\newblock Mllib: Machine learning in apache spark.
\newblock \emph{The Journal of Machine Learning Research}, 17(1):1235--1241.

\bibitem[{Miller(1998)}]{miller1998wordnet}
George~A Miller. 1998.
\newblock \emph{WordNet: An electronic lexical database}.
\newblock MIT press.

\bibitem[{Nikolentzos et~al.(2017)Nikolentzos, Meladianos, and
  Vazirgiannis}]{nikolentzos2017matching}
Giannis Nikolentzos, Polykarpos Meladianos, and Michalis Vazirgiannis. 2017.
\newblock Matching node embeddings for graph similarity.
\newblock In \emph{Thirty-First AAAI Conference on Artificial Intelligence}.

\bibitem[{Palmer et~al.(2004)Palmer, Babko-Malaya, and
  Dang}]{palmer2004different}
Martha Palmer, Olga Babko-Malaya, and Hoa~Trang Dang. 2004.
\newblock Different sense granularities for different applications.
\newblock In \emph{Proceedings of the 2nd International Workshop on Scalable
  Natural Language Understanding (ScaNaLU 2004) at HLT-NAACL 2004}, pages
  49--56.

\bibitem[{Palmer et~al.(2007)Palmer, Dang, and Fellbaum}]{palmer2007making}
Martha Palmer, Hoa~Trang Dang, and Christiane Fellbaum. 2007.
\newblock Making fine-grained and coarse-grained sense distinctions, both
  manually and automatically.
\newblock \emph{Natural Language Engineering}, 13(2):137--163.

\bibitem[{Pasini et~al.(2020)Pasini, Scozzafava, and
  Scarlini}]{pasini2020clubert}
Tommaso Pasini, Federico Scozzafava, and Bianca Scarlini. 2020.
\newblock Clubert: A cluster-based approach for learning sense distributions in
  multiple languages.
\newblock In \emph{Proceedings of the 58th Annual Meeting of the Association
  for Computational Linguistics}, pages 4008--4018.

\bibitem[{Peters et~al.(2018)Peters, Neumann, Iyyer, Gardner, Clark, Lee, and
  Zettlemoyer}]{peters2018deep}
Matthew~E Peters, Mark Neumann, Mohit Iyyer, Matt Gardner, Christopher Clark,
  Kenton Lee, and Luke Zettlemoyer. 2018.
\newblock Deep contextualized word representations.
\newblock \emph{arXiv preprint arXiv:1802.05365}.

\bibitem[{Pilehvar and Camacho-Collados(2018)}]{pilehvar2018wic}
Mohammad~Taher Pilehvar and Jose Camacho-Collados. 2018.
\newblock Wic: the word-in-context dataset for evaluating context-sensitive
  meaning representations.
\newblock \emph{arXiv preprint arXiv:1808.09121}.

\bibitem[{{R Core Team}(2018)}]{baseR}
{R Core Team}. 2018.
\newblock \emph{R: A Language and Environment for Statistical Computing}.
\newblock R Foundation for Statistical Computing, Vienna, Austria.

\bibitem[{Radford et~al.(2018)Radford, Narasimhan, Salimans, and
  Sutskever}]{radford2018improving}
Alec Radford, Karthik Narasimhan, Tim Salimans, and Ilya Sutskever. 2018.
\newblock Improving language understanding by generative pre-training.
\newblock \emph{URL https://s3-us-west-2. amazonaws.
  com/openai-assets/researchcovers/languageunsupervised/language understanding
  paper. pdf}.

\bibitem[{Rumshisky(2011)}]{rumshisky2011crowdsourcing}
Anna Rumshisky. 2011.
\newblock Crowdsourcing word sense definition.
\newblock In \emph{Proceedings of the 5th Linguistic Annotation Workshop},
  pages 74--81. Association for Computational Linguistics.

\bibitem[{Rumshisky et~al.(2012)Rumshisky, Botchan, Kushkuley, and
  Pustejovsky}]{rumshisky2012word}
Anna Rumshisky, Nick Botchan, Sophie Kushkuley, and James Pustejovsky. 2012.
\newblock Word sense inventories by non-experts.
\newblock In \emph{LREC}, pages 4055--4059.

\bibitem[{Siglidis et~al.(2018)Siglidis, Nikolentzos, Limnios, Giatsidis,
  Skianis, and Vazirgiannis}]{siglidis2018grakel}
Giannis Siglidis, Giannis Nikolentzos, Stratis Limnios, Christos Giatsidis,
  Konstantinos Skianis, and Michalis Vazirgiannis. 2018.
\newblock Grakel: A graph kernel library in python.
\newblock \emph{arXiv preprint arXiv:1806.02193}.

\bibitem[{Spearman(1904)}]{spearman1904proof}
Charles Spearman. 1904.
\newblock The proof and measurement of association between two things.

\bibitem[{Vaswani et~al.(2017)Vaswani, Shazeer, Parmar, Uszkoreit, Jones,
  Gomez, Kaiser, and Polosukhin}]{vaswani2017attention}
Ashish Vaswani, Noam Shazeer, Niki Parmar, Jakob Uszkoreit, Llion Jones,
  Aidan~N Gomez, Lukasz Kaiser, and Illia Polosukhin. 2017.
\newblock Attention is all you need.
\newblock In \emph{Advances in neural information processing systems}, pages
  5998--6008.

\bibitem[{Webber et~al.(2010)Webber, Moffat, and Zobel}]{webber2010similarity}
William Webber, Alistair Moffat, and Justin Zobel. 2010.
\newblock A similarity measure for indefinite rankings.
\newblock \emph{ACM Transactions on Information Systems (TOIS)}, 28(4):1--38.

\bibitem[{Weischedel et~al.(2011)Weischedel, Pradhan, Ramshaw, Palmer, Xue,
  Marcus, Taylor, Greenberg, Hovy, Belvin et~al.}]{weischedel2011ontonotes}
Ralph Weischedel, Sameer Pradhan, Lance Ramshaw, Martha Palmer, Nianwen Xue,
  Mitchell Marcus, Ann Taylor, Craig Greenberg, Eduard Hovy, Robert Belvin,
  et~al. 2011.
\newblock Ontonotes release 4.0.
\newblock \emph{LDC2011T03, Philadelphia, Penn.: Linguistic Data Consortium}.

\end{thebibliography}
\bibliographystyle{acl_natbib_nourl}

\end{document}